\documentclass[unnumsec,webpdf,contemporary,large]{mam-authoring-template}
\usepackage{helvet}

\newcommand{\seabornfont}{\fontfamily{phv}\selectfont}
\usepackage{hyperref}  
\usepackage{cleveref}
\usepackage{ulem} 
\usepackage{lmodern}

\crefname{figure}{Figure}{Figures}
\crefname{table}{Table}{Tables}
\crefname{algorithm}{Algorithm}{Algorithms}
\crefname{equation}{Eq.}{Equations}
\crefname{section}{Section}{Sections}
\creflabelformat{equation}{#2#1#3}

\usepackage{booktabs}      
\usepackage{multirow}
\usepackage{amsfonts}    
\usepackage{amsmath}
\usepackage{amssymb}
\usepackage{graphicx}
\usepackage[subrefformat=parens]{subfig}

\usepackage{pgfplots}
\usetikzlibrary{patterns}
\pgfplotsset{
    compat=1.8,
    tick align=inside,
    major tick length=2pt,
    legend style={font=\small\seabornfont, at={(0.5,1.02)}, anchor=north,legend columns=-1},
    tick label style={font=\small\seabornfont},
    axis line style=very thin,
    every axis/.append style={
        {font=\scriptsize\seabornfont},
    },
    label style={font=\small\seabornfont},
    width=8.8cm,
    height=6cm
}

\usepackage[dvipsnames]{xcolor}
\definecolor{DeepBlue}{HTML}{1F77B4}
\definecolor{OrangeRust}{HTML}{FF7F0E}
\definecolor{ForestGreen}{HTML}{2CA02C}
\definecolor{MaroonBrickRed}{HTML}{D62728}
\definecolor{Teal}{HTML}{17BECF}
\definecolor{MutedPurple}{HTML}{9467BD}
\definecolor{OliveGreen}{HTML}{8C564B}
\definecolor{GoldenYellow}{HTML}{BCBD22}
\definecolor{DarkGreyCharcoal}{HTML}{7F7F7F}
\definecolor{LightBlueSkyBlue}{HTML}{AEC7E8}
\definecolor{SlateGray}{HTML}{708090}
\definecolor{MustardYellow}{HTML}{FFDB58}
\definecolor{SeaGreen}{HTML}{2E8B57}
\definecolor{Lavender}{HTML}{C2C2FF}

\definecolor{green1}{rgb}{0.90, 0.96, 0.89}  
\definecolor{green2}{rgb}{0.75, 0.90, 0.76}  
\definecolor{green3}{rgb}{0.50, 0.78, 0.59}  
\definecolor{green4}{rgb}{0.30, 0.65, 0.40}  
\definecolor{green5}{rgb}{0.15, 0.50, 0.25}  
\definecolor{green6}{rgb}{0.10, 0.40, 0.18}  
\definecolor{green7}{rgb}{0.00, 0.27, 0.11}  

\definecolor{blue1}{rgb}{0.890, 0.949, 0.968}  
\definecolor{blue2}{rgb}{0.741, 0.862, 0.913}  
\definecolor{blue3}{rgb}{0.588, 0.779, 0.854}  
\definecolor{blue4}{rgb}{0.419, 0.682, 0.812}  
\definecolor{blue5}{rgb}{0.258, 0.572, 0.776}  
\definecolor{blue6}{rgb}{0.129, 0.443, 0.710}  
\definecolor{blue7}{rgb}{0.031, 0.250, 0.506}  

\definecolor{DarkRed}{rgb}{0.7, 0, 0}  

\definecolor{seagreen}{RGB}{46,139,87}

\pgfplotscreateplotcyclelist{mycolors}{
  {fill=blue!20,draw=black},
  {fill=blue!50,draw=black},
  {fill=blue!80,draw=black},
  {fill=seagreen!20,draw=black},
  {fill=seagreen!35,draw=black},
  {fill=seagreen!50,draw=black},
  {fill=seagreen!65,draw=black},
  {fill=seagreen!80,draw=black},
  {fill=seagreen!95,draw=black},
  {fill=seagreen!100,draw=black}
}

\graphicspath{{Fig/}}

\theoremstyle{thmstyleone}%
\theoremstyle{thmstyletwo}%
\theoremstyle{thmstylethree}%

\begin{document}

\journaltitle{Bioinformatics} 
\DOI{DOI}
\copyrightyear{2026}
\pubyear{2026}
\access{Advance Access Publication Date: Day Month Year}
\appnotes{Original Article}

\firstpage{1}

\makeatletter
\def\ps@titlepage{%
  \let\@oddfoot\@empty
  \let\@evenfoot\@empty
  \def\@oddhead{}
  \let\@evenhead\@oddhead
}
\makeatother


\title[IgPose: A Generative Data-Augmented Pipeline for Robust Immunoglobulin-Antigen Binding Prediction]{IgPose: A Generative Data-Augmented Pipeline for Robust Immunoglobulin-Antigen Binding Prediction}

\author[1,$\dagger$]{\textbf{Tien-Cuong Bui}} 
\author[1,$\dagger$]{\textbf{Injae Chung}} 
\author[1]{\textbf{Wonjun Lee}}
\author[1,$\ast$]{\textbf{Junsu Ko}} 
\author[1,2,3,$\ast$]{\textbf{Juyong Lee}} 

\authormark{Bui, Chung et al.}

\address[1]{\orgname{Arontier Co., Ltd.}, \orgaddress{\street{241 Gangnam-daero, Seocho-gu}, \postcode{06735}, \state{Seoul}, \country{Republic of Korea}}}
\address[2]{\orgdiv{Department of Molecular Medicine and Biopharmaceutical Sciences}, \orgname{Graduate School of Convergence Science and Technology, Seoul National University}, \orgaddress{\postcode{08826}, \state{Seoul}, \country{Republic of Korea}}}
\address[3]{\orgdiv{Research Institute of Pharmaceutical Science, College of Pharmacy}, \orgname{Seoul National University}, \orgaddress{\postcode{08826}, \state{Seoul}, \country{Republic of Korea}}}

\corresp[$\ast$]{Correspondence: \href{nicole23@snu.ac.kr}{nicole23@snu.ac.kr}, \href{junsuko@arontier.co}{junsuko@arontier.co}\\}
\corresp[$\dagger$]{These authors contributed equally to this work.} 


\abstract{
\vspace{3pt}
\noindent {\secsize\textcolor{jnlclr}{Motivation}}\\[1.5ex]
Predicting immunoglobulin-antigen (Ig-Ag) binding remains a significant challenge due to the paucity of experimentally-resolved complexes and the limited accuracy of \textit{de novo} Ig structure prediction. \\[4ex]
\noindent {\secsize\textcolor{jnlclr}{Results}}\\[1.5ex]
We introduce IgPose, a generalizable framework for Ig-Ag pose identification and scoring, built on a generative data-augmentation pipeline. 
To mitigate data scarcity, we constructed the Structural Immunoglobulin Decoy Database (SIDD), a comprehensive repository of high-fidelity synthetic decoys. 
IgPose integrates equivariant graph neural networks, ESM-2 embeddings, and gated recurrent units to synergistically capture both geometric and evolutionary features. 
We implemented interface-focused $k$-hop sampling with biologically guided pooling to enhance generalization across diverse interfaces. 
The framework comprises two sub-networks\textemdash IgPoseClassifier for binding pose discrimination and IgPoseScore for DockQ score estimation\textemdash and achieves robust performance on curated internal test sets and the CASP-16 benchmark compared to physics and deep learning baselines. 
IgPose serves as a versatile computational tool for high-throughput antibody discovery pipelines by providing accurate pose filtering and ranking.\\[4ex]
\noindent {\secsize\textcolor{jnlclr}{Availability and Implementation}}\\[1.5ex]
IgPose is available on GitHub (https://github.com/arontier/igpose).\\[4ex]
\noindent {\secsize\textcolor{jnlclr}{Contact}}\\[1.5ex]
Juyong Lee (nicole23@snu.ac.kr), Junsu Ko (junsuko@arontier.co).
}
\maketitle

\section{Introduction}

Immunoglobulin-antigen (Ig-Ag) recognition is a cornerstone of adaptive immunity \citep{lu2018antibodyreview} and underpins the development of many therapeutic biologics \citep{chan2025antibodyreview}, molecular diagnostics \citep{garcia2021covidtest}, and vaccine development \citep{crank2019vaccine}. 
Accurate structure modeling of Ig-Ag complexes, identifying correct spatial orientations and binding poses, and estimating binding affinities are critical for the rational design of therapeutic antibodies \citep{norman2019antibodydesign}. 
However, this task remains challenging because of the limited number of experimentally resolved Ig-Ag structures, the conformational plasticity of complementarity-determining region (CDR) loops, and the high-dimensional, heterogeneous nature of protein-protein interfaces (PPI). Collectively, these factors hinder the out-of-distribution generalization of existing computational models, often resulting in the failure of conventional interface scoring functions applied to novel epitope landscapes.

Traditional physics-based tools such as Rosetta \citep{alford2017rosetta} and Prodigy \citep{xue2016prodigy} offer interpretable energy-based scoring terms for PPI but frequently underperform when distinguishing near-native from non-native Ig-Ag binding poses at scale. Recent deep learning (DL) methods for protein-protein pose classification and scoring, such as TRScore \citep{guo2022trscore}, GNN-DOVE \citep{wang2021protein}, DeepRank-GNN-ESM \citep{xu2024deeprank}, and ProAffinity-GNN \citep{zhou2024proaffinity}, improve scoring by learning from interface representations, but typically lack geometric equivariance and often exhibit overfitting and/or reduced performance on unseen structures. 
Recently, an antibody-specific B cell epitope prediction tool, AbEpiTope-1.0 \citep{clifford2025abepitope}, was proposed to employ AlphaFold-2 Multimer (AFM) \citep{evans2021protein} for structural modeling and ESM-IF embeddings \citep{hsu2022learning} pooled over Ig-Ag interface nodes with a shallow multi-layer perceptron (MLP) to discriminate native-like Ig-Ag poses and estimate interface quality. 
However, varying performance across different benchmarks suggests their limited generalizability. 

\begin{figure*}[htp]
  \centering
  \includegraphics[width=0.95\linewidth]{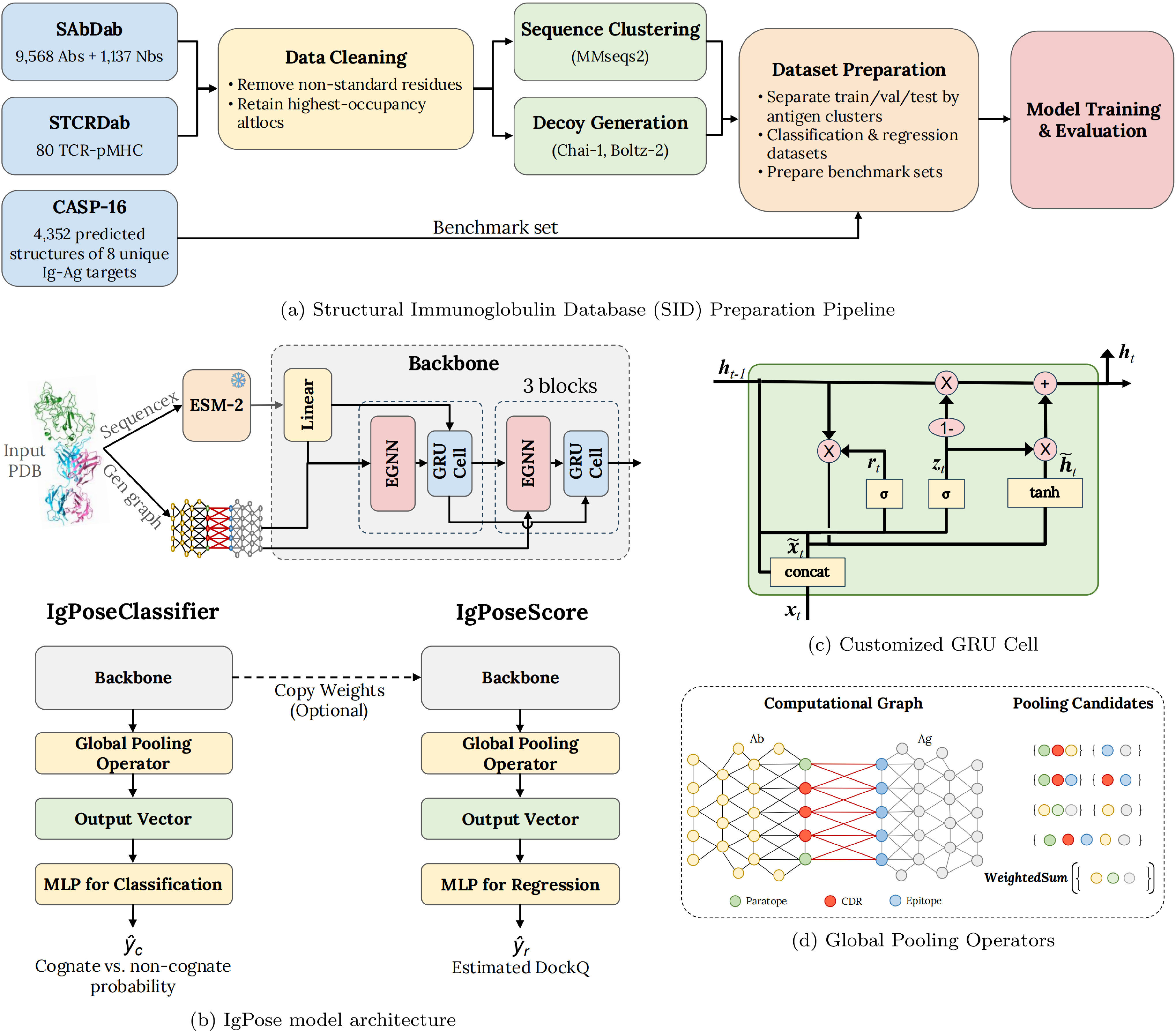}%
  \caption{An overview of IgPose model architecture and its internal components. (\textbf{a}) The data preparation pipeline from data collection, data cleaning, and decoy generation with Chai-1 \citep{chaidiscovery2024} and Boltz-2 \citep{wohlwend2024boltz1,passaro2025boltz} to data preparation and model training and evaluation. (\textbf{b}) The IgPose - equivariant message-passing architecture built upon EGNN \citep{satorras2021n} and a customized GRU module. (\textbf{c}) The architecture of the customized GRU module. (\textbf{d}) Alternative global pooling operators corresponding to different selective regions.}
  \label{fig:overview}
\end{figure*}

Here, we introduce IgPose, an Ig-Ag complex scoring model based on a generative framework that integrates evolutionary context, geometric inductive bias, structural augmentation, and task-specific learning objectives (\textbf{\cref{fig:overview}}).
IgPose enriches contextual features with evolutionary knowledge from the ESM-2 embeddings \citep{lin2023evolutionary} of Ig and Ag sequences and employs E(n)-equivariant graph neural networks (EGNN) \citep{satorras2021n} coupled with a customized gated recurrent unit (GRU) module \citep{cho2014properties} to model long-range interactions while preserving physical symmetries. 
To overcome the scarcity of experimentally determined Ig-Ag structures in public repositories, we introduced a generative pipeline to supplement the training set with various synthetic decoy structures modeled using Chai-1 \citep{chaidiscovery2024} and Boltz-2 \citep{wohlwend2024boltz1,passaro2025boltz}. 
Our framework employs two core models, IgPoseClassifier and IgPoseScore, to address both Ig-Ag pose classification and regression tasks. 
IgPoseClassifier performs binary classification of a given Ig–Ag structure into native-like or non-native conformations.
IgPoseScore estimates the DockQ score \citep{basu2016dockq} of a complex model structure. 
We design task-specific loss functions for each tool to facilitate efficient model training. 

Our evaluation results demonstrate that IgPose consistently outperforms traditional physics-based methods \citep{alford2017rosetta,xue2016prodigy} and existing DL models \citep{wang2021protein,guo2022trscore,zhou2024proaffinity} across various benchmark sets. 
Inconsistent performance trends of baseline methods emphasize their limitations in generalizability under distribution shifts and structural variations. 
In contrast, IgPose achieves up to a two-fold improvement in Area Under the Precision-Recall Curve (AP) scores across diverse benchmarks, which we attribute to the integration of our generative data-augmentation strategy with a hybrid architecture combining EGNNs, evolutionary ESM-2 embeddings, and GRUs to capture complex interfacial features. These results position IgPose as a robust computational framework for antibody discovery, facilitating more accurate discrimination and ranking of lead candidates during virtual screening.

\section{Experimental Methods} \label{sec:method}
\subsection{Problem Formulation} \label {problem_formulation}

Let \(S\) denote an Ig-Ag complex provided in PDB format, containing antibody chain(s) and one antigen chain. We represent \(S\) as a computational graph $\mathcal{G} = \bigl(\mathcal{V},\,\mathcal{E},\,\mathcal{X}_v,\,\mathcal{X}_e,\,\mathcal{P}\bigr)$, where \(\mathcal{V} = \{v_i\}_{i=1}^N\) is a set of \(N\) amino‐acid residues, $\mathcal{E}$ is a set of connections between them, and \(\mathcal{P} = \{p_a \in \mathbb{R}^3 \mid a \in \bigcup_{i=1}^N \mathrm{atoms}(v_i)\}\) collects the 3D positions of all atoms. We seek two functions. 

First, a classifier $f_{\theta}$ is defined as follows:
\[
f_{\theta}: \mathcal{G}\to[0,1], \quad \hat y_c = f_{\theta}(\mathcal{G}),
\]
where \(\hat y_c\) is the probability that \(S\) is a native-like binding pose of a cognate Ig-Ag pair (\(y=1\)) versus a non-native decoy (\(y=0\)) including a non-native bound pose of the cognate Ig-Ag pair or any bound pose of a non-cognate Ig-Ag pair.
Second, a regressor $r_{\phi}$ is defined as follows:
\[
r_{\phi}: \mathcal{G}\to\mathbb{R}, \quad \hat y_r = r_{\phi}(\mathcal{G}),
\]
where \(\hat y_r\) is an estimated DockQ score of a given model complex structure.

\subsection{Data Sources and Curation}
We assembled the \textit{Structural Immunoglobulin Database} (SID), a comprehensive structural dataset of Ig-Ag complexes by combining experimentally determined structures from public repositories with systematically generated decoys.

\medskip
\noindent \textbf{Structural Antibody Database (SAbDab):}
X-ray crystallogra\-phic (XRC) and electron cryomicroscopy (cryo-EM) structures of antibodies (Ab; 4,362 structures) and nanobodies (Nb; 1,137 structures) bound to \emph{monomeric} (single chain) antigens, deposited prior to September 23, 2024, were curated from SAbDab \citep{dunbar2014sabdab}. 
During data cleaning, non-standard amino acids with \emph{HETATM} records were removed. When residues had alternative locations (\emph{altlocs}), the one with the highest occupancy was retained.
To generate training clusters, antigen sequences\textemdash derived from \emph{SEQRES} records or the UniProt database \citep{uniprot2025}\textemdash were clustered by 30\% sequence similarity and 30\% sequence coverage using MMseqs2 \citep{steinegger2017,steinegger2018}, yielding 942 clusters comprising 5,499 Ig-Ag pairs.

\medskip
\noindent \textbf{Structural T-Cell Receptor Database (STCRDab):}
A curated set of high-resolution, annotated structures of T-cell receptors (TCR) bound to peptide-MHC (major histocompatibility complex; pMHC) complexes from STCRDab \citep{leem2018stcrdab} were fetched from \citet{zhang2023framedipt}.
Structures were similarly cleaned to remove non-standard amino acids and low occupancy \emph{altlocs}. A total of 80 TCR-pMHC complexes were clustered by MHC class (I or II).

\medskip
\noindent \textbf{Structural Immunoglobulin Decoy Dataset (SIDD):}
Based on the structural data curated from SAbDab and STCRDab, we constructed SIDD, an \textit{in-house} database of decoy Ig-Ag structures generated using biomolecular structure prediction tools, Chai-1 \citep{chaidiscovery2024} and Boltz-2 \citep{wohlwend2024boltz1,passaro2025boltz}.
\begin{enumerate}
  \item \emph{Cognate Ig-Ag decoys:} an \textit{in silico} generated dataset composed of re-predicted structures of cognate Ab-Ag, Nb-Ag, and TCR-pMHC pairs, yielding \textit{ca.} \(9.2\times 10^3\) structures. 
  A DockQ score of 0.8 (\textit{high-quality}) \citep{basu2016dockq} was used as a threshold to discriminate between \emph{positive} ($\geq 0.8$) and \emph{negative} ($< 0.8$) decoy structures.
  \item \emph{Non-cognate Ig-Ag decoys:} an \textit{in silico} generated dataset composed of Abs, Nbs, or TCRs in complex with non-cognate antigens. 
  Ig and Ag sequences were clustered by 100 and 90\% similarity, respectively, and 80\% coverage using MMseqs2 \citep{steinegger2017,steinegger2018}. One sample was selected per cluster to remove redundancy.
  To avoid potential artifacts arising from matching antigens (i.e., false positives), we purposefully paired Abs and Nbs with pMHCs, and TCRs with monomeric antigens, generating \textit{ca.} \(10^5\) \emph{negative} decoy structures. 
\end{enumerate}
The generated structures were clustered following the same sequence-based clustering procedure used for the SAbDab and STCRDab datasets, based on their corresponding target antigen (monomeric antigen or pMHC).

\subsection{Benchmark Dataset}
\noindent \textbf{Critical Assessment of Structure Prediction 16 (CASP-16):}
We collated all structure predictions submitted for eight Ig-Ag docking targets released at the CASP-16 competition \citep{CASP16} (predictioncenter.org/casp16): \emph{H1204} (PDB-8vyl), \emph{H1215} (unreleased), \emph{H1222} (PDB-9cqd), \emph{H1223} (PDB-9cqb), \emph{H1225} (PDB-9cqa), \emph{H1232} (PDB-9cn2), \emph{H1233} (PDB-9cbn), and \emph{H1244} (unreleased). 
These complexes, which include Abs, Nbs, and single-chain variable fragments (scFvs) bound to their cognate antigens, are experimentally resolved but unpublished (at the time of the competition), ensuring that their 3D structures are novel and `unseen' by existing computational prediction methods, including IgPose. To ensure there was no data leakage between our training cut-off (2024.09.23) and the CASP-16 modeling end date (2024.08.31), we cross-checked all eight CASP-16 targets against our curated dataset and confirmed that none were present.
During data cleaning, each Ig copy in a structure was paired with its bound antigen(s). 
Similar to SIDD, we used a DockQ score \citep{basu2016dockq} of 0.8 to discriminate between \emph{positive} and \emph{negative} decoy structures; DockQ scores were provided by the CASP-16 assessors \citep{CASP16}. The CASP-16 dataset is composed of 4,352 predicted structures.

\subsection{Dataset Grouping}
We grouped \textit{SID} into three internal datasets for classification and regression tasks. 
\textbf{\cref{tab:data_ratio}} presents statistics for these datasets and the CASP-16 benchmark. 
The first classification dataset (SID-CA) comprises all experimentally determined `native' Ig-Ag structures drawn from SAbDab and STCRDab, along with cognate and non-cognate conformations generated by Chai-1 \citep{chaidiscovery2024}.
To mitigate bias from antigen homology, samples were stratified by antigen clusters into training, validation, and test splits in an approximate 6:2:2 ratio. 
To investigate generalization across alternative structure prediction tools, we designated a second classification test set (SID-CB) for structures generated exclusively with Boltz-2 \citep{wohlwend2024boltz1,passaro2025boltz}. 

To train a regression model estimating DockQ scores of cognate decoys, we constructed a regression dataset (SID-R) which integrates experimentally-resolved structures from SAbDab and STCRDab, and Chai-1- and Boltz-2-generated decoy structures in SIDD.
Each structure is annotated with a continuous DockQ label (SAbDab, STCRDab = 1; non‑cognate decoys = 0; cognate decoys $ \in (0,1) $). To maintain an unbiased mix of \emph{negative} samples, we randomly sampled seven Chai-1 decoys per Ig chain within each antigen cluster. 
SID-R was then split by the same cluster IDs used in SID-C into training, validation, and test subsets.

Finally, we used the CASP‑16 Ig-Ag prediction results for external validation of both tasks. 
In the classification setting, we adopted the same DockQ threshold of 0.8 for \emph{positive}/\emph{negative} labeling to ensure that only the highest‑quality docking poses are labeled as \emph{positive}. 
This stringent cut-off guards against cases where strong signals on non‑Ig-Ag interfaces can inflate the overall (global) DockQ score, thus excluding conformations with suboptimal Ig-Ag contacts. 
In the regression setting, we instead predicted continuous DockQ values directly against the scores obtained by the assessors of CASP-16.

\begin{table}[!hb]
    \centering
    \caption{Statistics of our Structural Immunoglobulin Database (SID) and CASP-16 benchmark dataset. 
    Here, \emph{positive} represents conformations that are either experimentally-resolved `native' complexes or cognate decoys with DockQ scores $\geq 0.8$; \emph{negative} indicates structural decoys with DockQ scores $ < 0.8 $. 
    SID-C and SID-R denote datasets for classification and regression tasks, respectively. 
    SID-CA and SID-CB are two classification datasets containing decoy conformations generated by Chai-1 and Boltz-2, respectively.}
    \begin{tabular}{c|c|c|c}
        \toprule
         Dataset & Split & \#Positive    & \#Negative \\
         \midrule
         \multirow{3}{*}{SID-CA} & Train        & 3668  & 68907  \\
                                       & Validation   & 1445  & 9133       \\
                                       & Test         & 928   & 17092      \\
        \hline
        SID-CB                         & Test         & 1001   & 16436     \\
        \hline
        \multirow{3}{*}{SID-R }         & Train        & 4668  & 10599      \\
                                       & Validation   & 1921  & 1448       \\
                                       & Test         & 1025   & 1116      \\
         \hline
         CASP-16         & Benchmark    & 349   & 4003       \\
         \bottomrule
    \end{tabular}
    \label{tab:data_ratio}
\end{table}

\subsection{CDR annotation}
ANARCI \citep{dunbar2015anarci} was used to annotate and extract information about complementarity determining regions (CDRs) from Ab, Nb, scFv, and TCR sequences. 
The Chothia numbering scheme \citep{chothia1987chothia,al-lazikani1997chothia} was used to annotate Abs, Nbs, and scFvs, and the IMGT numbering scheme \citep{lefranc1999imgt,lefranc2003imgt,lefranc2015imgt} for TCRs.

\subsection{Graph Construction and Features}
Ig-Ag structures in PDB or mmCIF formats were converted into PDBQT files using MGLtools \citep{morris2009mgltools}. 
During this process, implicit (non-polar) hydrogens were removed, and explicit (polar) hydrogens were introduced where absent. 
For graph construction, we defined edge sets as follows:
$ \mathcal{E}_r = \{ (i,j) \mid \min_{a\in\mathcal{A}_i,\;b\in\mathcal{A}_j} \| p_a-p_b \|_2\le\tau_r  \},  $
where $\mathcal{A}$ is the set of atoms of the corresponding residue, $p \in\mathbb{R}^3$ denotes the Cartesian coordinate, and $r \in\{\mathrm{intra},\mathrm{inter}\}$. 
Specifically, $\mathcal{E}_{\mathrm{intra}}$ is the intra-residue edge set including connections between nodes within antibody/antigen residues s.t. $\tau_\mathrm{intra} \leq 3.5 \mathrm{\AA}$, while $\mathcal{E}_{\mathrm{inter}}$ includes inter-molecular edges at the binding interface s.t $\tau_\mathrm{inter} \leq 10.0  \mathrm{\AA} $. 

Input node features $\mathcal{X}_v\in\mathbb{R}^{N\times d_x}$ were obtained from ESM-2 \citep{lin2023evolutionary}, where $d_x = 320$. Input edge attributes $\mathcal{X}_e\in\mathbb{R}^{N\times d_e}$ comprise three raw distances (minimum atomic distance, C\(\alpha\)-C\(\alpha\) distance, and center-of-mass distance), expanded via a Gaussian RBF map $\phi_D:\mathbb{R}\!\to\!\mathbb{R}^D$ with $D$ log-spaced scales in $[0.25, 8]$, yielding a $d_e = 3 \times D$. We used $D = 10$ by default and varied $D$ in ablations.

\subsection{Model Architecture}

Here, we present our equivariant graph neural network (EGNN) architecture (\cref{fig:overview}).
Let the input graph be
$\mathcal{G}=(\mathcal{V},\mathcal{E},\mathcal{X}_v,\mathcal{X}_e,\mathcal{P})$. The network comprises $T$ EGNN \citep{satorras2021n} layers, each interleaved with a custom GRU gate (\cref{fig:overview}c), followed by a weighted pooling head and an MLP classifier. 
We map node features $X\in\mathbb{R}^{N\times d_x}$ into a hidden space:

\begin{equation}
H^{(0)} = \textrm{SiLU} \bigl(X W_{\mathrm{in}} + b_{\mathrm{in}}\bigr)\in\mathbb{R}^{N\times d_h},
\label{eq:linear}
\end{equation}

\noindent For $t=1,\dots,T$, we apply:
\begin{equation}
    \begin{aligned}
    \widetilde{H}^{(t)}, \mathcal{P}^{(t)} &= \mathrm{EGNN}\bigl(H^{(t-1)},\,\mathcal{P}^{(t-1)}\bigr),\ \\
    H^{(t)} &= \mathrm{GRU}\bigl(\bigl[\widetilde{H}^{(t)}\;|\;H^{(t-1)}\bigr],\,H^{(t-1)}\bigr).
    \end{aligned}
    \label{eq:egnn}
\end{equation}

Integrating the GRU function \citep{xiong2019pushing, zhou2024proaffinity} into GNNs is prevalent as it acts as a learnable gated residual update. As $H$ is an invariant feature in EGNN \citep{satorras2021n}, adding a gated connection on top does not break the equivariant properties of $\mathcal{P}$. In our design, the GRU function introduces an additional residual connection to ease the oversmoothing problem of GNNs by concatenating $H^{(t-1)}$ and $\widetilde{H}^{(t)}$ as the input state. This modification enforces a stronger self-loop to preserve a node's identity in message passing steps. It also introduces a direct linear path for gradients to flow backward to $H^{(t-1)}$ via the candidate gate $n$ even when other gates saturate. Further detail can be found in the \textbf{Supplementary Information}.

After $T$ equivariant message-passing iterations on $\mathcal{G}$, we perform a read-out function on a selected subset $\mathcal{S}\in\mathcal{V}$. The perturbation approach, exploring a set of essential node embeddings maximizing GNN performance, is common practice in explainable artificial intelligence (XAI) research \citep{bui2023generating, bui2023toward}. In practice, we can perform this procedure either through sampling-based and learning-based algorithms or with domain expertise. Given the enormous size of computational graphs and specificity of the Ig-Ag binding problem, we opt for the latter approach and introduce several strategies to define $\mathcal{S}$: \emph{All nodes} - pooling all nodes, \emph{Interface Only} - pooling only binding-interface nodes, \emph{CDR-Epitope Only} - pooling only nodes in edges between CDRs and epitopes, \emph{CDR Only} - pooling only nodes in CDRs, \emph{w/o interface} - pooling all nodes except those in inter Ig-Ag edges, \emph{w/o CDR-Epitope} - pooling all nodes except those in edges between CDRs and epitopes, \emph{w/o CDR} - pooling all nodes excluding those in CDRs (\cref{fig:overview}d). Node weights and the global graph embedding are then:
\begin{equation}
  w_i = \mathrm{sigmoid}\bigl(w_p^\top h_i + b_p\bigr),
  \quad g = \sum_{i\in\mathcal{S}} w_i\,h_i,
  \label{eq:read_out}
\end{equation}
with $|\mathcal{S}|\le N$, where $h_i$ corresponds to a row $i$ in $H^{(T)}$.

Our architecture supports both classification and regression tasks. 
We train classification models by minimizing the classification objective function on discrete labels. 
After that, we initialize a regression model from a trained classifier by replacing the softmax output with a linear layer and fine-tuning all parameters on continuous labels using a regression loss function. 
This transfer learning strategy improves regression performance with fewer data samples.
For the readout layers, the following layers are used.

\noindent IgPoseClassifier\textemdash the classifier\textemdash uses a two-layer MLP with SiLU activation and softmax:

\begin{equation}
\hat y_c = \mathrm{Softmax}\bigl(W^{(2)}_c \, \mathrm{SiLU}(W^{(1)}_c\,g + b^{(1)}_c) + b^{(2)}_c\bigr).
\end{equation}

\noindent IgPoseScore\textemdash the regressor\textemdash estimates DockQ scores and substitutes the classifier head with a scaled-tanh predictor:

\begin{equation}
    \hat y_r = \tfrac{1}{2}\bigl(\tanh(0.5 \times z) + 1\bigr),
    \quad z = W_r\,g + b_r.
\end{equation}

As deep ensembles \citep{lakshminarayanan2017simple} can improve predictive uncertainty estimation and robustness, we define a weighted ensemble function to aggregate the predicted results of $M$ models into $\hat{y}^*$, as follows:

\begin{equation}
\hat{y}^* = \sum_{m=1}^{M} w_m \, \hat{y},
\qquad 
\text{where } \sum_{m=1}^{M} w_m = 1,\; w_m \ge 0.
\end{equation}

\subsection{Training Objective Functions} 

We optimize specific objectives for each task. For classification, we learn $\theta$ by minimizing:
\begin{align}
  \theta^* &= \arg\min_\theta \mathcal{L}_c(\theta),\\
  \mathcal{L}_c &= \mathcal{L}_{\mathrm{GC}} + \alpha\,\mathcal{L}_{\mathrm{NC}} + \beta\,\mathcal{L}_{\mathrm{MDN}} .
\end{align}
Here, $\mathcal{L}_{\mathrm{GC}}$ is the graph-level cross-entropy loss that predicts the class of an input graph $\mathcal{G}$, $\mathcal{L}_{\mathrm{NC}}$ the node-level cross-entropy that predicts individual node types (e.g., antigen, heavy chain, light chain), and $\mathcal{L}_{\mathrm{MDN}}$ the negative Pearson correlation between predicted probability and ground-truth DockQ. The Lagrange multipliers set as $\alpha=10^{-3}$ and $\beta=2\times10^{-3}$. For regression, we optimize:
\begin{align}
  \phi^* &= \arg\min_\phi \mathcal{L}_r(\phi),\\
  \mathcal{L}_r &= \mathcal{L}_{\mathrm{coeff}} + \mathcal{L}_{\mathrm{rank}}.
\end{align}
Here, $\mathcal{L}_{\mathrm{coeff}} = - \mathrm{Corr}(y,\hat y_r)$ for maximizing Pearson correlation \citep{shen2023generalized}, and $\mathcal{L}_{\mathrm{rank}}$ is the listwise ranking loss \citep{cao2007learning}.

\subsection{Training and Evaluation Protocol}
\noindent \textbf{Baselines:} We benchmarked our models against MIEnsembles (the highest computed AUC, AP and $r$ scores on CASP-16 \citep{CASP16}), standard physics-based scoring functions (Prodigy \citep{xue2016prodigy} and Rosetta \citep{alford2017rosetta}) and publicly available DL models (TRScore \citep{guo2022trscore}, GNN‐DOVE \citep{wang2021protein}, DeepRank‐GNN‐ESM (DR-GNN-ESM) \citep{xu2024deeprank}, ProAffinity‐GNN \citep{zhou2024proaffinity}, and AbEpiTope-1.0 \citep{clifford2025abepitope}). All DL models are executed using their default configurations.

\noindent \textbf{Evaluation Metrics:}
We independently evaluated the classification and regression performance. For classification, we report five standard metrics: \emph{Precision} (P), \emph{Recall} (R), \emph{F1-score} (F1), \emph{the Area Under the Receiver Operating Characteristic Curve} (AUC), and and \emph{the Area Under the Precision-Recall Curve} (AP).

For regression, we quantify the agreement between the predicted values $\hat{y}_{i}$ and the true targets $y_{i}$ using the Pearson correlation coefficient ($r$).

\medskip
\noindent \textbf{IgPose Implementations:} Processing an entire graph \(\mathcal{G}\) is suboptimal in both prediction performance and computational efficiency. 
Therefore, we extracted a subgraph \(\mathcal{G}_s\) via 3-hop sampling (\textbf{\cref{alg:khop_sampling}}) around interface edges \(\mathcal{E}_{\mathrm{inter}}\), restricting message-passing to relevant binding-site neighborhoods. 
Specifically, we implemented two alternative sampling strategies: starting from unique nodes in $\mathcal{E}_\mathrm{inter}$ or from nodes in CDRs, with both procedures stopping once $\mathcal{G}_s$ reached  a pre-defined threshold of 600. 
Node embeddings \(x_i\in\mathbb{R}^{320}\) were derived from the ESM-2 (8M parameter) model \citep{lin2023evolutionary} as larger variants caused severe overfitting. The default edge embedding dimension was 30 ($D = 10$) and varied in ablation studies.

In addition to EGNN \citep{satorras2021n}-based IgPose models, we also implemented two variants and trained them with the SID-CA dataset. First, we substituted EGNN \citep{satorras2021n} with FastEGNN \citep{zhang2025fast} and performed weighted sum pooling over all nodes. The second model includes a two-layer high-order equivariant message passing network (MACE \citep{batatia2022mace}) followed by a simple sum pooling layer and an MLP classification head.

Our models, implemented in PyTorch and DGL \citep{wang2019dgl}, comprised of four EGNN layers, each followed by a custom GRU gate, with a hidden dimension \(h=64\). 
The final classifier has an intermediate dropout \(p=0.1\).  
Training ran up to 50 epochs with early stopping based on validation F1 scores. 
Adam optimizer was used with an initial learning rate \(lr=10^{-4}\) and a cosine annealing scheduler scaling $lr$ down to $10^{-5}$. 
To address class imbalance, a weighted random sampler was employed, setting weights for negative and positive samples at 0.8 and 0.2, respectively.

\section{Results} \label{sec:results}

\subsection{Classification Performance} 

We evaluated IgPose on three test sets: SID-CA, SID-CB, and CASP‑16. For a fair comparison, we ran all baselines on our evaluation set and defined optimal cut-off thresholds based on their respective F-beta scores (\textbf{\cref{alg:f1_threshold}}). We also performed a weighted ensemble of IgPoseClassifier and AbEpiTarget.

As shown in \textbf{\cref{fig:performance}a} and \textbf{\cref{tab:detailed_classification}}, IgPose outperforms both physics‑ and DL-based baselines on AUC and AP scores, showing its strong discriminatory power across all generation protocols: Chai-1 (SID-CA), Boltz-2 (SID-CB), or mixed (CASP-16).
Physics‑based methods behave inconsistently: Rosetta performs well on SID‑C but drops to modest accuracy on CASP‑16, while Prodigy underperforms on the SID‑C test set and shows only marginal improvement on CASP‑16.
All general-purpose DL models demonstrate poor performance across the three tasks, suggesting poor generalization.
MIEnsembles, a top EMA method from \citep{CASP16}, achieves a slightly better AP score on CASP-16 than both energy-based and general DL-based methods.
Interestingly, AbEpiTarget outperforms IgPoseClassifier on CASP‑16 despite its limited performance on the SID-C test set, likely due to a bias toward its AlphaFold-Multimer--generated data \citep{evans2021protein,clifford2025abepitope}. 
Among our variants, FastEGNN \citep{zhang2025fast} and MACE \citep{batatia2022mace} only achieve high AUC and AP scores on our internal SID-C tests, while the ensemble of IgPoseClassifier and AbEpiTarget (IgPC-AbET) consistently outperforms all methods.

\begin{table*}[t]
\caption{Detailed classification performance comparison of our three implemented models (FastEGNN, MACE, IgPoseClassifier) and baselines on two internal test sets and CASP-16 on Precision (P), Recall (R), F1, AUC-ROC (AUC), and AUC-PR (AP) scores. Baseline methods produce unique results for each dataset, IgPoseClassifier's results are from its deployed version. In IgPC-AbET setting, we perform weighted average on output probabilities of IgPoseClassifier and AbEpiTarget with weights of 0.7 and 0.3, respectively. Bold and underlined text represent the best and second best scores of a metric accordingly.}
\centering
\resizebox{\textwidth}{!}{%
\begin{tabular}{c|ccccc|ccccc|ccccc}
\toprule
\multirow{2}{*}{\textbf{Method}} &
\multicolumn{5}{c|}{\textbf{SID-CA}} &
\multicolumn{5}{c|}{\textbf{SID-CB}} &
\multicolumn{5}{c}{\textbf{CASP-16}} \\
 & P & R & F1 & AUC & AP & P & R & F1 & AUC & AP & P & R & F1 & AUC & AP \\
\midrule

\multicolumn{1}{l|}{MIEnsembles} & - & - & - & - & - & - & - & - & - & - & \textbf{-} & - & \textbf{-} & 0.894 & 0.326 \\
\multicolumn{1}{l|}{Prodigy} & 0.008 & \textbf{1.000} & 0.015 & 0.071 & 0.007 & 0.012 & \textbf{1.000} & 0.024 & 0.093 & 0.021 & 0.080 & \textbf{1.000} & 0.148 & 0.677 & 0.129 \\
\multicolumn{1}{l|}{Rosetta} & 0.598 & 0.871 & \underline{0.709} & 0.959 & \underline{0.862} & 0.435 & 0.857 & 0.577 & 0.939 & 0.823 & 0.127 & \underline{0.983} & 0.225 & 0.884 & 0.268 \\
\multicolumn{1}{l|}{TRScore} & 0.249 & 0.108 & 0.150 & 0.623 & 0.124 & 0.281 & 0.113 & 0.161 & 0.632 & 0.138 & 0.241 & 0.192 & 0.214 & 0.781 & 0.189 \\
\multicolumn{1}{l|}{GNN-DOVE} & 0.040 & 0.156 & 0.063 & 0.380 & 0.040 & 0.735 & 0.136 & 0.229 & 0.507 & 0.239 & 0.143 & 0.980 & 0.250 & 0.811 & 0.174 \\
\multicolumn{1}{l|}{ProAffinityGNN} & 0.055 & 0.232 & 0.089 & 0.625 & 0.100 & 0.092 & 0.233 & 0.132 & 0.785 & 0.131 & 0.131 & 0.418 & 0.200 & 0.573 & 0.095 \\
\multicolumn{1}{l|}{DR-GNN-ESM} & 0.037 & 0.270 & 0.065 & 0.414 & 0.041 & 0.074 & 0.263 & 0.116 & 0.513 & 0.128 & 0.117 & 0.921 & 0.208 & 0.565 & 0.083 \\
\multicolumn{1}{l|}{AbEpiTarget} & 0.293 & 0.120 & 0.170 & 0.705 & 0.152 & 0.217 & 0.135 & 0.167 & 0.621 & 0.125 & \textbf{0.591} & 0.854 & \textbf{0.699} & \textbf{0.965} & \underline{0.880} \\
\hline
\multicolumn{1}{l|}{IgPoseClassifier} & \textbf{0.940} & 0.490 & 0.644 & 0.982 & 0.888 & \textbf{0.967} & 0.466 & 0.628 & \underline{0.981} & 0.917 & 0.474 & 0.900 & 0.621 & 0.914 & 0.747 \\
\multicolumn{1}{l|}{IgPC-AbET} & \textbf{0.940} & 0.477 & 0.633 & 0.987 & 0.888 & \underline{0.964} & 0.476 & \underline{0.637} & \textbf{0.990} & \textbf{0.945} & \underline{0.568} & 0.897 & \underline{0.696} & \underline{0.928} & \textbf{0.896} \\

\multicolumn{1}{l|}{MACE} & 0.898 & \underline{0.895} & \textbf{0.897} & 0.972 & \textbf{0.920} & 0.828 & \underline{0.873} & \textbf{0.850} & 0.961 & \underline{0.894} & 0.176 & 0.914 & 0.295 & 0.888 & 0.459 \\
\multicolumn{1}{l|}{FastEGNN} & \underline{0.923} & 0.362 & 0.520 & \underline{0.982} & 0.840 & 0.927 & 0.343 & 0.500 & \underline{0.981} & 0.875 & 0.000 & 0.000 & 0.000 & 0.813 & 0.207 \\

\bottomrule
\end{tabular}%
}
\label{tab:detailed_classification}
\end{table*}
\input{fig_cls_reg}

\begin{figure*}[ht]
  \centering
  \includegraphics[width=0.9\textwidth]{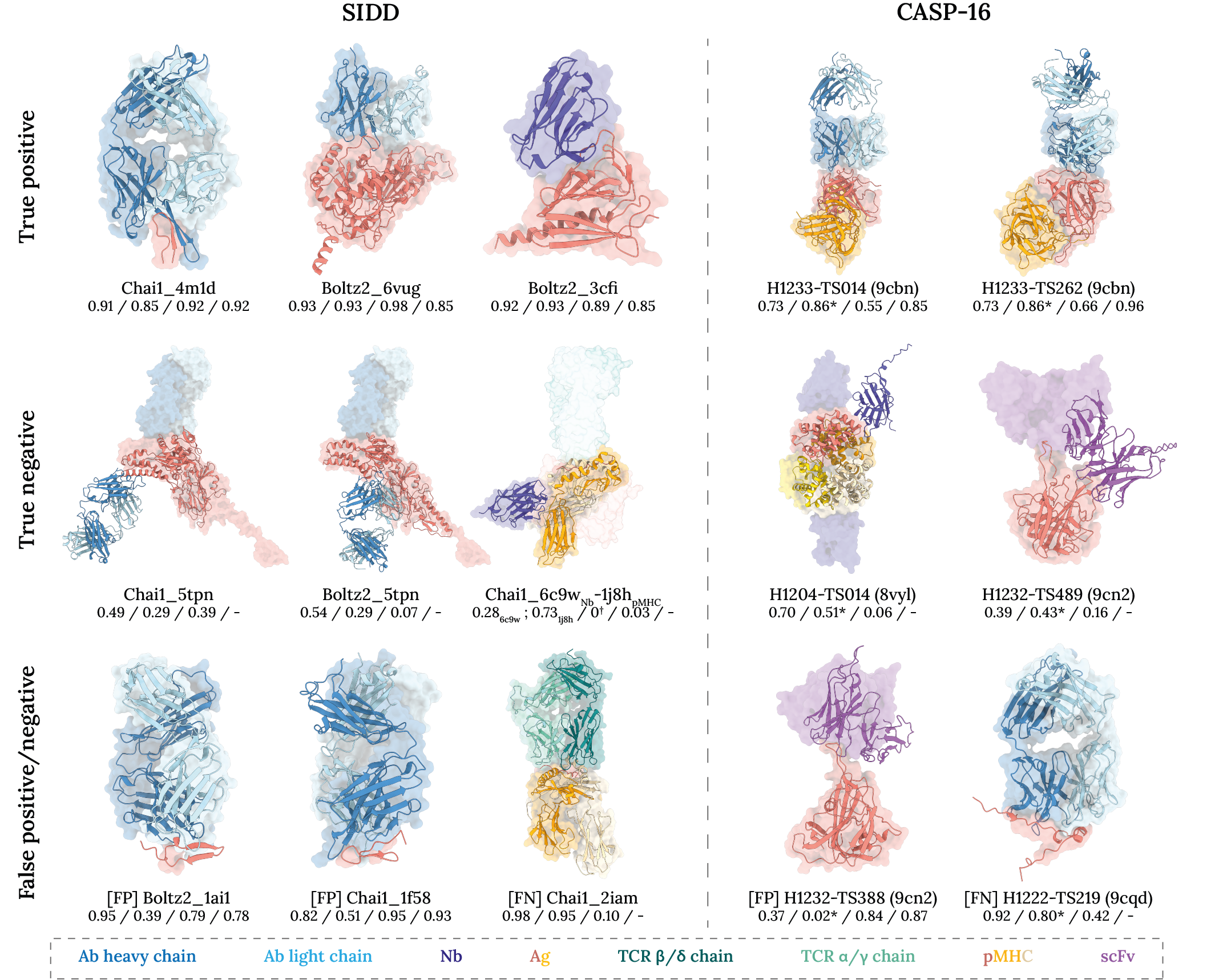}%
  \caption{Representative decoy structures from our SIDD test dataset and CASP-16 competition submissions. 
  The decoy structures -- shown as cartoon -- are categorized into four groups based on the IgPoseClassifier prediction: true positive, true negative, false positive, and false negative. Ground truth structures are shown as transparent surface.
  For SIDD dataset structures, the prefixes `Chai1' and `Boltz2' specify the computational method used for decoy generation, followed by letters that reference the original PDB-ID. 
  For CASP-16 structures \citep{CASP16}, `H12[number]' denotes the CASP-16 target ID and `TS[number]' corresponds to the participating group that submitted the prediction. PDB-IDs of ground-truth structures are shown in parenthesis. 
    Each structure is accompanied by four numerical scores: (from left to right) global TMscore, global DockQ score, IgPoseClassifier score, and IgPoseScore. 
  IgPoseScore was computed for true/false positive structures only. Asterisk (*) indicates DockQ scores provided by CASP-16 assessors; dagger ($\dagger$) indicates incomputable DockQ scores, which apply exclusively to non-cognate Ig-Ag decoys. 
  Color-coding scheme used to distinguish antibodies (Ab), nanobodies (Nb), antigens (Ag), T-cell Receptors (TCR), peptide-MHC complexes (pMHC), and single-chain variable fragments (scFv) is indicated at the bottom of the figure.}
  \label{fig:example}
\end{figure*}

In \textbf{\cref{fig:example}}, we illustrate representative decoy structures from SIDD and prediction submissions in CASP-16, categorized into TP, TN, FP, and FN according to IgPoseClassifier predictions. 
These examples encompass structures generated using established structure prediction tools such as Chai-1 \citep{chaidiscovery2024} and Boltz-2 \citep{wohlwend2024boltz1,passaro2025boltz}, along with various approaches employed by independent groups in the CASP-16 competition \citep{CASP16}, which collectively capture the structural diversity of binding interfaces, chain compositions, and epitope-paratope arrangements. 
Each structure is annotated with TM-score \citep{zhang2022usalign}, DockQ \citep{basu2016dockq}, IgPoseClassifier probabilities, and IgPoseScore, enabling quantitative evaluation. 
These results demonstrate IgPoseClassifier's ability to identify native-like interfaces, while also highlighting its challenges with specific decoys. 

\subsection{Regression Performance}

We next evaluated the binding quality scoring capability of IgPoseScore against the baseline methods (\textbf{\cref{fig:performance}b}).
Each baseline employed distinct scoring modalities to rank Ig–Ag poses: estimated TM-score in MIEnsembles \citep{CASP16}, binding energies in Rosetta \citep{alford2017rosetta} and Prodigy \citep{xue2016prodigy}, estimated IoU with crystal structures in AbEpiScore \citep{clifford2025abepitope}, $\mathrm{p}K_{d}$ in ProAffinityGNN \citep{zhou2024proaffinity}, probabilities in TRScore \citep{guo2022trscore}, GNN-Dove \citep{wang2021protein}, and $F_\mathrm{nat}$ in DeepRank-GNN-ESM \citep{xu2024deeprank}. 
For IgPoseScore, we tested two variants: FS (trained from scratch) and FT (finetuned from IgPoseClassifier by replacing the head with regression). We also performed a weighted ensemble of IgPoseScore and AbEpiScore (IgPS-AbES).

As shown in \textbf{\cref{fig:performance}b}, IgPoseScore exhibits the highest $r$ score on the internal SID-R test set ($r=0.653$), with the fine-tuned version (IgPoseScore-FT) providing additional gains. 
While MACE performs comparably to IgPoseScore on SID-R ($r=0.634$), it underperforms on CASP-16 ($r=0.233$).
Rosetta and Prodigy also show inconsistent performance across the two datasets. Specifically, Rosetta strongly correlates with DockQ on SID-R ($r=0.551$) but shows little to no correlation on CASP-16. Conversely, Prodigy achieves a surprisingly high correlation score ($r=0.3812$) on CASP-16 but a negative correlation on SID-R. FastEGNN and MIEnsembles demonstrate moderate performance, with FastEGNN obtaining $r=0.440$ and $r=0.263$ on CASP-16 and SID-R, respectively, and MIEnsembles achieving  $r=0.232$ on the CASP-16 dataset ($r$ for SID-R is not available).
Interestingly, most deep learning baselines (DeepRank‑GNN‑ESM, TRScore, GNN‑DOVE, and ProAffinityGNN) show near-zero or negative correlations. 
Finally, although AbEpiScore is comparable to IgPoseScore on CASP‑16 ($r=0.3114$ vs. $r=0.355$), it falls behind our model on the SID-R test set ($r=0.321$ vs. $r=0.653$).
However, the weighted ensemble of these two models (IgPS-AbES) achieves the highest $r$ scores on the two test sets (0.686 on SID-R and 0.415 on CASP-16).

\subsection{Candidate Selection Performance} 

Practical use cases require accurate ranking of predicted binders to prioritize antibody candidates for experimental validation. 
We evaluated candidate selection performance by first filtering out predicted non-binders using IgPoseClassifier, AbEpiTarget, and their ensemble IgPC-AbET, then ranking the remaining samples with IgPoseScore, AbEpiScore, and their ensemble IgPS-AbES. Top-K success rates (precision@K) were calculated on SID-R and CASP-16 test sets.

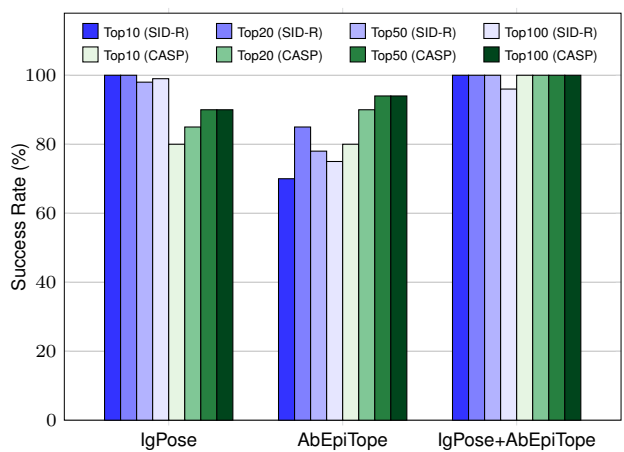
\begin{figure}[t]
    \centering
    \begin{tikzpicture}
  \begin{axis}[
      width=8.9cm,
      height=7cm,
      ybar,
      bar width=6pt,
      enlarge x limits=0.3,
      enlarge y limits={upper,value=0.18},
      symbolic x coords={IgPose,AbEpiTope, IgPose+AbEpiTope},
      xtick=data,
       ymin=0,
    ymax=100,
    ymajorgrids=true,
    ylabel={Success Rate (\%)},
    ylabel shift=-0.25cm,
    tick label style={font=\small\seabornfont},
    legend style={
        at={(0.5,0.85)},anchor=south,
        legend columns=4,
        /tikz/every even column/.append style={column sep=8pt},
        /tikz/every odd column/.append style={column sep=0pt},
        draw=none,
        font=\tiny\seabornfont
    },
    legend image post style={
        xshift=0pt,  
    },
    legend cell align=left,
    legend image code/.code={\draw[fill=#1] (0.2cm, 0.13cm) rectangle ( 0cm, -0.07cm);},
    ]


    \addplot+[bar shift=-21pt,fill=blue!80,draw=black] coordinates {(IgPose,100) (AbEpiTope,70) (IgPose+AbEpiTope,100)};
    \addplot+[bar shift=-15pt,fill=blue!50,draw=black] coordinates {(IgPose,100) (AbEpiTope,85) (IgPose+AbEpiTope,100)};
    \addplot+[bar shift=-9pt, fill=blue!30,draw=black] coordinates {(IgPose,98)  (AbEpiTope,78) (IgPose+AbEpiTope,100)};
    \addplot+[bar shift=-3pt, fill=blue!10,draw=black] coordinates {(IgPose,99)  (AbEpiTope,75) (IgPose+AbEpiTope,96)};
    
    
    \addplot+[bar shift=+3pt, fill=green1,draw=black] coordinates {(IgPose,80) (AbEpiTope,80) (IgPose+AbEpiTope,100)};
    \addplot+[bar shift=+9pt, fill=green3,draw=black] coordinates {(IgPose,85)  (AbEpiTope,90) (IgPose+AbEpiTope,100)};
    \addplot+[bar shift=+15pt,fill=green5,draw=black] coordinates {(IgPose,90)  (AbEpiTope,94) (IgPose+AbEpiTope,100)};
    \addplot+[bar shift=+21pt,fill=green7,draw=black] coordinates {(IgPose,90)  (AbEpiTope,94) (IgPose+AbEpiTope,100)};

    \legend{
      Top10 (SID-R),Top20 (SID-R),Top50 (SID-R),Top100 (SID-R),
      Top10 (CASP),Top20 (CASP),Top50 (CASP),Top100 (CASP)
    }
  \end{axis}
\end{tikzpicture}

    \caption{Comparison of Top-K success rates for IgPose and AbEpiTope on two benchmark datasets. Blue-shaded and green-shaded bars show the SID-R and CASP-16 success rates, respectively. Here, success rate is defined as the precision in Top-K: the proportion of true positive samples among the Top-K ranked predictions. Here, IgPoseClassifier or AbEpiTarget are first used to filter out predicted negative samples; the remaining predicted positives are then ranked by IgPoseScore or AbEpiScore; precision is then calculated among the top 10, 20, 50, and 100 ranked candidates. \emph{IgPose+AbEpiTope} denotes the weighted ensemble variant of the two methods.}
    \label{fig:ranking_candidate}
\end{figure}

As shown in \textbf{\cref{fig:ranking_candidate}}, IgPose outperforms AbEpiTope on the SID-R test set, achieving near-perfect success rates: 100\% at Top-10 and Top-20, 98\% at Top-50, and 99\% at Top-100. 
AbEpiTope-1.0 shows slightly lower precision across all thresholds: 70\% at Top-10, 85\% at Top-20, 78\% at Top-50, and 75\% at Top-100. 
On CASP-16, both models achieve 80\% at Top-10.
AbEpiTope performs marginally better than IgPose at Top-20, Top-50, and Top-100 (90\%, 94\%, 94\% vs. 85\%, 90\%, 90\%).
Notably, the ensemble variant outperforms both individual models, achieving nearly perfect scores across all Top-K metrics.

\subsection{Ablation Study}
We performed several ablation studies to dissect the contribution of each framework component. The results show that: (1) selective global pooling over broader, non-contacting regions improves robustness on structurally diverse decoys, whereas restricting pooling to CDR/interface regions limits generalization (\textbf{\cref{fig:pooling_methods}}); (2) graph-level augmentations ($k$-hop sampling) stabilize generalization while embedding perturbations degrade it (\textbf{\cref{fig:data_augmentation}}); (3) all-atom graphs clearly outperform $C_\alpha$ graphs, and large edge embeddings harm generalization (\textbf{\cref{fig:graph_construction}}); (4) ensemble methods increase robustness and mitigates overfitting (\textbf{\cref{fig:all-prob-dists}}); (5) a customized GRU cell accelerates early learning and yields consistent AP score gains, particularly on CASP-16  (\textbf{\cref{fig:test_set_vs_casp16}}). 
Further information can be found in the \textbf{Supplementary Information}.

\section{Discussion}  \label{sec:discussion}

Modeling Ig–Ag interactions remains challenging, as neither conventional physics‑based tools nor DL-based models reliably distinguish true from false binding poses. 
While general-purpose PPI baselines utilize large-scale training sets such as the  Protein-Protein Docking Benchmark (ZDOCK) \citep{hwang2010zdockbm4,vreven2015zdockbm5}, DockGround \citep{liu2008dockground,collins2022dockground}, or PDBbind \citep{wang2004pdbbind} (with cut-offs typically preceding 2022), they often lack the specialized inductive biases \citep{battaglia2018relational} required for the highly flexible CDR loops found in antibody–antigen interfaces. Furthermore, some general-purpose DL models, such as DeepRank-GNN-ESM \citep{xu2024deeprank}, explicitly exclude antibody data from their training sets. This lack of domain-specific data curation and training likely accounts for their subpar performance on CASP-16 targets compared to the antibody-specific architecture of IgPose.
These shortcomings necessitate a unified framework that can filter plausible docking decoys with high precision and estimate Ig-Ag binding quality (e.g. DockQ \citep{basu2016dockq}) to prioritize candidates for experimental validation.

We developed IgPose, a generative framework that augments limited experimental Ig-Ag complexes with diverse structural decoys generated using Chai-1 \citep{chaidiscovery2024} and Boltz-2 \citep{wohlwend2024boltz1,passaro2025boltz}. IgPose integrates both geometric and evolutionary information to model Ig-Ag binding: it combines equivariant message passing with interface-focused subgraph sampling to capture spatial patterns invariant under physical transformations, while incorporating evolutionary information from ESM-2 protein language model embeddings \citep{lin2023evolutionary}. IgPose also introduces various pooling strategies to let the neural network prioritize critical information autonomously. Crucially, IgPose provides two complementary tools for evaluating antibody candidates: (i) IgPoseClassifier enables high-confidence discrimination of cognate from non-cognate binding poses, offering an effective filter for large-scale docking pipelines; (ii) IgPoseScore predicts binding pose scores, supporting finer-grained prioritization of antibody candidates. 

Empirical evaluations showed that IgPose generalizes effectively across datasets, highlighting the benefits of generative data augmentation and geometry-aware modeling \textbf{\cref{fig:performance}b}. We observed that conventional binding energy estimation tools are significantly more sensitive to the `physicality' of atomic level structural geometries. 
Rosetta, for instance, likely achieved superior performance on our internal SID test sets by strictly penalizing the unphysical atomic interactions and steric clashes often inherent in hallucinated binding contacts. However, its discriminative power likely diminished on the CASP-16 benchmark due to most submitted structures undergoing extensive conformational refinement (e.g. massive global sampling, relaxation, molecular dynamics refinement), which results in a narrow distribution of Rosetta scores (\textbf{\cref{fig:rosetta_prodigy_score}}), making it difficult to distinguish between a near-native low-energy pose and an energetically minimized but incorrect decoy.
In contrast, Prodigy assigns favorable (more negative) binding affinities to structures with a high number of interfacial contacts. This heuristic makes it susceptible to hallucinated interactions in decoys, resulting in an inverse predictive pattern in our internal tests. On CASP-16, Prodigy achieved moderate AUC and $r$ scores, reflecting its capacity to distinguish tight interfaces from loose contacts. However, it failed to discriminate high-quality complexes ($\text{DockQ} \ge 0.8$) from acceptable- and medium-quality poses, resulting in a low AP score. 
General DL-based prediction models also suffered from out-of-distribution problems, leading to inferior performances across benchmarks. These findings suggest that existing tools remain unreliable for ranking in-silico Ig-Ag structures.
Notably, AbEpiTope \citep{clifford2025abepitope}, a recent Ab-Ag binding prediction framework, achieved substantial performance on CASP-16 but performed poorly on our internal SID dataset. This performance gap suggests that the model's efficacy may be sensitive to specific data augmentation strategies and training objectives, and integrating such approaches into our framework is a promising avenue for future investigation.

The Ig–Ag binding prediction field currently lacks standardized, large-scale benchmarks. While the CASP challenge \citep{CASP16} provides an invaluable external evaluation set, it is constrained by a limited number of cognate Ig-Ag pairs (n = 8). The lack of highly accurate Ig-Ag structure modeling tools and limited benchmarks hinder objective comparisons between studies. Although our generative data pipeline partially addresses this scarcity, the curated dataset may introduce a bias toward antibody structures encountered during training, potentially impeding generalization through partial memorization. To mitigate this, we employed an ensemble approach \citep{lakshminarayanan2017simple}. We observed that a simple average of multiple trained checkpoints consistently enhanced IgPose's performance across all benchmarks. Furthermore, a weighted ensemble of IgPose and AbEpiTope outperformed all comparison methods, further confirming the effectiveness of this synergistic approach. Interestingly, our evaluation of pooling strategies suggests that the model does not prioritize the binding interface in its readout functions, despite this region containing the most biologically informative signals. This discrepancy may arise from data imbalance or inductive biases \citep{battaglia2018relational} in the architecture, necessitating further investigation into the spatial distribution of attention during training.

From a practical perspective, IgPose is well-suited for therapeutic antibody discovery. In a typical virtual screening workflow, IgPoseClassifier can rapidly discard non-cognate Ig-Ag structures, while IgPoseScore facilitates candidate prioritization through pose scoring. This two-step inference framework optimizes resource allocation for downstream wet-lab validation, particularly when screening against structurally novel antigens or expansive antibody libraries. However, several challenges remain: (i) IgPose's Pearson correlation with DockQ remains moderate, and (ii) while IgPose demonstrates high accuracy for antibody-antigen binding prediction, we observed comparatively lower performance on the TCR-pMHC subset (\textbf{\cref{fig:target_fixed}}). This disparity highlights the distinct biophysical challenges of TCR recognition, such as the lower binding affinity ranges and more constrained binding orientations compared to antibodies, which will be the focus of future model refinements.

\section{Conclusion} \label{sec:conclusion}

Existing methods for Ig–Ag pose discrimination and scoring frequently fail to generalize across diverse antigen landscapes.
To address this, we developed IgPose, a unified framework designed to enhance the robustness and generalizability of Ig–Ag pose classification and scoring. 
IgPose achieves this by synergistically integrating generative decoy augmentation, evolutionary context via ESM-2 embeddings, and geometric inductive biases through equivariant graph neural networks (EGNNs) with gated updates. Furthermore, the inclusion of auxiliary losses strategically aligns structural uncertainty with predictive confidence.

Our framework incorporates IgPoseClassifier for identifying near-native cognate poses and IgPoseScore for scoring poses.
Compared to traditional physics- and DL-based tools, the two functions of IgPose demonstrated superior classification accuracy and regression correlation, showcasing IgPose's consistent generalization to unseen structures.
Additionally, detailed ablation studies confirmed that interface-focused $k$-hop subgraph sampling and selective global pooling over specific regions are pivotal to the performance of the model. 
These specific designs align the model's inductive biases with the complex topology of Ig-Ag structures and interfaces, significantly enhancing discriminative power.
We envision IgPose as a practical tool for therapeutic antibody discovery pipelines, where fast and accurate screening of binding poses and affinities can significantly accelerate candidate selection.

\section{Data availability}
The Ig-Ag structures used in this study were obtained from public repositories: SAbDab \citep{dunbar2014sabdab}, STCRDab \citep{leem2018stcrdab}, and CASP-16 \citep{CASP16}. 
The data underlying this article are available on Zenodo (https://doi.org/10.5281/zenodo.17431183). 

\section{Code availability}
The source code for IgPose is available on Zenodo \\(https://doi.org/10.5281/zenodo.17431131) and Github \\(https://github.com/arontier/igpose).

\section{Competing interests}
All authors are employees of Arontier Co., Ltd.

\section{Author contributions}
T-C.B. designed and implemented models, conducted experiments, analyzed results, and wrote the manuscript.
I.C. collated and curated Ig-Ag data (SIDD), processed structural data, provided biological insight and directions, analyzed and interpreted the results, and contributed to manuscript writing.  
W.L. contributed to SIDD curation, data processing, and initial environment setup.
J.K. conceived the research plan, contributed to data collection and manuscript writing.
J.L. conceived the research plan, co-designed the model, provided critical feedback, analyzed and interpreted the results, and contributed to manuscript writing. 
T-C.B. and I.C. agreed that the order of their respective names may be changed for personal pursuits to best suit their own interests.

\section{Acknowledgments}
This research was supported by the Bio \& Medical Technology Development Program of the National Research Foundation (NRF) funded by the Korean government (MSIT) (Grants NRF-2022M3E5F3081268 and NRF-2022M3A9J4079198). J.L. was supported by Seoul National University (370C-20220109 and AI-Bio Research Grant 0413-20230053), and the National Research Foundation of Korea (Grants 2020M3A9G7103933, RS-2023-00256320, and 2022R1C1C1005080)

\bibliographystyle{unsrtnat}
\bibliography{reference}

\clearpage

\setcounter{section}{0}
\renewcommand{\thesection}{S\arabic{section}}
\setcounter{figure}{0}
\renewcommand{\thefigure}{S\arabic{figure}}
\setcounter{table}{0}
\renewcommand{\thetable}{S\arabic{table}}
\setcounter{page}{1}

\clearpage
\twocolumn[
  \begin{center}
    {\large \bfseries \textcolor{jnlclr}{Supplementary Information for:} \\ [0.5em]
    \Large {\textcolor{jnlclr}{IgPose: A Generative Data-Augmented Pipeline for Robust Immunoglobulin-Antigen Binding Prediction}} \par}
    
    \vspace{1.5em}
    
    {\large \bfseries Tien-Cuong Bui$^{1,\dagger}$, Injae Chung$^{1,\dagger}$, Wonjun Lee$^1$, Junsu Ko$^{1,\ast}$, and Juyong Lee$^{1,2,\ast}$ \par}
    
    \vspace{1em}
    {\small $^1$Arontier Co., Ltd., Seoul, 06735, Republic of Korea \\
    $^2$Seoul National University, Seoul, 08826, Republic of Korea \par}
    
    \vspace{1em}
    {\small $^\dagger$These authors contributed equally to this work. \\
    $^\ast$Correspondence: nicole23@snu.ac.kr, junsuko@arontier.co}
    \vspace{2em} 
  \end{center}
]

\section{Additional Experiments}
\subsection{Pooling Strategies} \label{sec:pooling_strategy}

This section describes global pooling strategies presented in \textbf{\cref{fig:pooling_methods}}. A detailed comparison is shown in \textbf{\cref{tab:pooling_strategy}}.

\begin{itemize}
    \item \textbf{Pooling over all nodes}: The default configuration in our architecture for the global pooling operator is to perform weighted sum over all nodes in an input graph.
    \item \textbf{Pooling over interface (\emph{interface-only})}: Nodes located at the interface region are selected for global pooling operator.
    \item \textbf{Pooling over CDR (\emph{CDR-only})}: Nodes located at the CDR region are selected for global pooling operator.
    \item \textbf{Pooling over CDR-Epitope (\emph{CDR-Epitope-only})}: Nodes located at the CDR and epitope regions are selected for global pooling operator.
    \item \textbf{Excluding interface (\emph{w/o interface})}: Nodes located at the interface region are selected for global pooling operator.
    \item \textbf{Excluding CDR (\emph{w/o CDR})}: Nodes located at the CDR region are selected for global pooling operator.
    \item \textbf{Excluding CDR-Epitope (\emph{w/o CDR-Epitope})}: Nodes located at the CDR and epitope regions are selected for global pooling operator.
    \item \textbf{Ensemble of 3 best}: Predicted probabilities of three best pooling techniques are averaged and selected as the final predictions.
\end{itemize}

\subsection{Selective Global Pooling Strategies} 

\begin{table*}[!hb]
\caption{Detailed classification performance comparison of pooling strategies on SID-CA test and CASP-16 on Precision (P), Recall (R), F1, AUC-ROC (AUC), and AUC-PR (AP) scores. Bold and underlined text represent the best and second best scores of a metric accordingly. Best checkpoints are selected based on AP scores on the evaluation set.}
\centering
\small
\begin{tabular}{c|ccccc|ccccc}
\toprule
\multirow{2}{*}{\textbf{Strategy}} &
\multicolumn{5}{c|}{\textbf{Our Test Set}} &
\multicolumn{5}{c}{\textbf{CASP-16}} \\
 & P & R & F1 & AUC & AP & P & R & F1 & AUC & AP \\
 
\midrule
\multicolumn{1}{l|}{Pooling over all nodes}  & 0.866 & 0.723 & \textbf{0.788} & \underline{0.968} & 0.846 & 0.353 & \underline{0.900} & 0.507 & 0.891 & 0.352 \\
\hline
\multicolumn{1}{l|}{w/o interface nodes}    & 0.627 & \textbf{0.868} & 0.728 & 0.959 & 0.756 & 0.157 & \textbf{0.908} & 0.268 & 0.902 & 0.525 \\
\multicolumn{1}{l|}{w/o CDR nodes}   & 0.598 & \underline{0.824} & 0.693 & 0.959 & 0.713 & 0.192 & \textbf{0.908} & 0.317 & \textbf{0.916} & 0.529 \\
\multicolumn{1}{l|}{w/o CDR-Epitope nodes}   & 0.954 & 0.334 & 0.494 & 0.966 & 0.849 & \textbf{0.653} & 0.868 & \textbf{0.745} & 0.910 & 0.682 \\
\hline
\multicolumn{1}{l|}{interface-only nodes} & 0.675 & 0.802 & 0.733 & 0.958 & 0.694 & 0.181 & 0.894 & 0.301 & 0.776 & 0.158 \\
\multicolumn{1}{l|}{CDR-only nodes} & 0.814 & 0.255 & 0.388 & 0.899 & 0.518 & 0.001 & 0.003 & 0.002 & 0.462 & 0.071 \\
\multicolumn{1}{l|}{CDR-Epitope-only nodes} & 0.901 & 0.671 & \underline{0.769} & 0.955 & 0.830 & 0.000 & 0.000 & 0.000 & 0.639 & 0.099 \\
\hline
\multicolumn{1}{l|}{Ensemble of 3 best checkpoints}    & \underline{0.932} & 0.502 & 0.653 & 0.967 & \underline{0.854} & \underline{0.400} & 0.897 & \underline{0.553} & \underline{0.915} & \textbf{0.759} \\
\multicolumn{1}{l|}{Ensemble of 5 best checkpoints}    & \textbf{0.934} & 0.519 & 0.668 & \textbf{0.969} & \textbf{0.869} & 0.392 & \underline{0.900} & 0.546 & \underline{0.915} & \underline{0.755} \\

\bottomrule
\end{tabular}
\label{tab:pooling_strategy}

\vspace{2em}

\caption{Detailed classification performance comparison of data augmentation methods on SID-CA test and CASP-16 on Precision (P), Recall (R), F1, AUC-ROC (AUC), and AUC-PR (AP) scores. Weighted sum pooling over all nodes are applied to all settings. Bold and underlined text represent the best and second best scores of a metric accordingly.}
\centering
\small
\begin{tabular}{c|ccccc|ccccc}
\toprule
\multirow{2}{*}{\textbf{Method}} &
\multicolumn{5}{c|}{\textbf{Our Test Set}} &
\multicolumn{5}{c}{\textbf{CASP-16}} \\
 & P & R & F1 & AUC & AP & P & R & F1 & AUC & AP \\
 
\midrule
\multicolumn{1}{l|}{Nondocking} & \textbf{0.920} & 0.586 & \underline{0.716} & \textbf{0.973} & \underline{0.835} & 0.171 & \textbf{0.911} & 0.288 & 0.875 & \underline{0.329} \\
\multicolumn{1}{l|}{Random rotation} & 0.457 & \underline{0.781} & 0.576 & 0.937 & 0.352 & 0.156 & \underline{0.908} & 0.266 & 0.854 & 0.245 \\
\multicolumn{1}{l|}{Mask Embed} & 0.617 & 0.529 & 0.570 & 0.945 & 0.528 & \underline{0.338} & 0.897 & \underline{0.491} & 0.858 & 0.254 \\
\multicolumn{1}{l|}{Mask Embed + random rotation} & 0.800 & 0.516 & 0.628 & \textbf{0.973} & 0.817 & 0.132 & 0.622 & 0.218 & 0.649 & 0.102 \\
\multicolumn{1}{l|}{3-hop sampling with node threshold} & \underline{0.866} & 0.723 & \textbf{0.788} & \underline{0.968} & \textbf{0.846} & \textbf{0.353} & 0.900 & \textbf{0.507} & \textbf{0.891} & \textbf{0.352} \\
\multicolumn{1}{l|}{CDR sampling with node threshold} & 0.601 & \textbf{0.828} & 0.696 & 0.957 & 0.776 & 0.168 & \underline{0.908} & 0.284 & \underline{0.883} & 0.326 \\
\bottomrule
\end{tabular}
\label{tab:data_augmentation}

\end{table*}

To determine the structural information essential for accurate predictions, we evaluated the impact of global pooling operations on model performance. 
We employed two strategies: one that excludes the Ig-Ag interface, CDR region, and CDR-epitope interface, and another that used only these specific regions. We also assessed two additional configurations: an ensemble that averaged predictions from the three best models and a baseline that applied weighted sum pooling across all nodes.

\begin{figure}[ht]
\centering
\begin{tikzpicture}
\begin{axis}[
    width=8.9cm,
    height=7cm,
    ybar,
    bar width=8pt,
    enlarge x limits=0.6,
    enlarge y limits={upper,value=0.05},
    ylabel={AP Score},
    ylabel shift=-0.15cm,
    symbolic x coords={SID-CA Test,CASP-16},
    xtick=data,
    ymin=0,
    ymax=1.0,
    ymajorgrids=true,
    tick label style={font=\small\seabornfont},
    legend style={
        at={(0.5,0.85)},anchor=south,
        legend columns=4,
        /tikz/every even column/.append style={column sep=5pt},
        /tikz/every odd column/.append style={column sep=0pt},
        draw=none,
        font=\tiny\seabornfont
    },
    legend image post style={
        xshift=0pt,  
    },
    legend cell align=left,
    legend image code/.code={\draw[fill=#1] (0.2cm, 0.13cm) rectangle ( 0cm, -0.07cm);},
]

\addplot [bar shift=-28pt,fill=blue!80,draw=black] coordinates {
  (SID-CA Test,0.854)
  (CASP-16,0.759)
};

\addplot [bar shift=-20pt,fill=blue!50,draw=black] coordinates {
  (SID-CA Test,0.756)
  (CASP-16,0.525)
};

\addplot [bar shift=-12pt,fill=blue!30,draw=black] coordinates {
  (SID-CA Test,0.713)
  (CASP-16,0.529)
};

\addplot [bar shift=-4pt,fill=blue!10,draw=black] coordinates {
  (SID-CA Test,0.849)
  (CASP-16,0.682)
};

\addplot [bar shift=4pt,fill=green1,draw=black] coordinates {
  (SID-CA Test,0.846)
  (CASP-16,0.352)
};

\addplot [bar shift=12pt,fill=green3,draw=black] coordinates {
  (SID-CA Test,0.694)
  (CASP-16,0.158)
};

\addplot [bar shift=20pt,fill=green5,draw=black] coordinates {
  (SID-CA Test,0.518)
  (CASP-16,0.071)
};

\addplot [bar shift=28pt,fill=green7,draw=black] coordinates {
  (SID-CA Test,0.830)
  (CASP-16,0.099)
};

\legend{
Ensemble of 3 best,
w/o interface,
w/o CDR,
w/o CDR–Epitope,
All nodes,
Interface-only,
CDR-only,
CDR–Epitope-only
}

\end{axis}
\end{tikzpicture}
\caption{A comparison of performance of pooling strategies on SID-CA test and CASP‑16 benchmarks. Labels denote the set of nodes used in the global weighted sum pooling operation. `w/o' denotes exclusion of the specified node set from pooling, while the suffix `only' indicate exclusive use of the specified set of nodes. \emph{All node}: weighted sum pooling over all nodes in a graph. \emph{Ensemble of 3 best}: average of output probabilities from the three models corresponding to the pale blue bars. Further detail can be found in \textbf{\cref{sec:pooling_strategy}}.}
\label{fig:pooling_methods}
\end{figure}
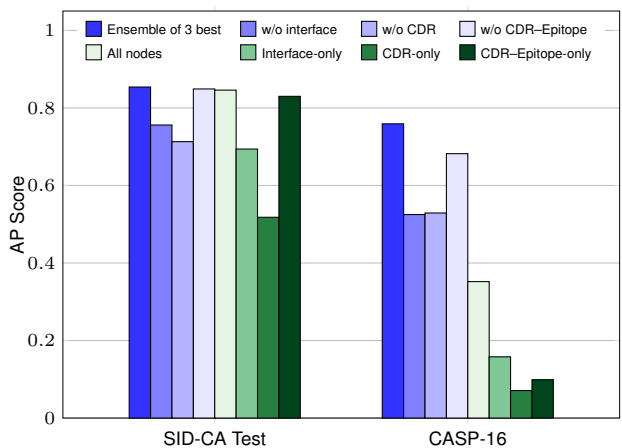

Pooling strategies yield similarly high AP scores on SID‑CA, suggesting that the choice of pooling region has limited impact on this curated internal dataset (\textbf{\cref{fig:pooling_methods}}), which may be due to partial memorization of structural patterns during training. The CASP‑16 set \citep{CASP16} inherently has a greater structural diversity reflecting a broad spectrum of computational prediction and refinement protocols employed by various participants. On this unseen dataset, broader pooling strategies that incorporate non-contacting scaffold regions significantly enhance model robustness by capturing the global topological features of the Ig–Ag complex.
Specifically, pooling strategies that focus on the framework while excluding the immediate interface or CDR loops performed substantially better than global pooling over all nodes. This suggests that non-contacting regions of the immunoglobulin and antigen provide a critical geometric reference frame, effectively `anchoring' the binding site within the overall three-dimensional structure.
Conversely, restricted pooling over only the CDR or CDR–epitope regions leads to a sharp degradation in performance, demonstrating that the hypervariable CDR loops alone lack sufficient geometric or contextual information for out-of-distribution generalization. Their high conformational plasticity and sequence variability likely render local-only signals too `noisy' to reliably rank poses without the stabilizing context provided by the conserved protein scaffold.

\subsection{Data Augmentation Methods} \label{sec:data_augmentation}

\begin{figure}[ht]
\centering
\begin{tikzpicture}
\begin{axis}[
    width=8.9cm,
    height=7cm,
    ybar,
    bar width=8pt,
    enlarge x limits=0.6,
    enlarge y limits={upper,value=0.0},
    ylabel={AP Score},
    ylabel shift=-0.15cm,
    symbolic x coords={SID-CA Test,CASP-16},
    xtick=data,
    ytick={0,0.2,0.4,0.6,0.8,1},
    ymin=0,
    ymax=1.0,
    ymajorgrids=true,
    tick label style={font=\small\seabornfont},
    legend style={
        at={(0.7,0.8)},
        anchor=south,
        legend columns=2,
        /tikz/every even column/.append style={column sep=10pt},
        /tikz/every odd column/.append style={column sep=0pt},
        draw=none,
        font=\tiny\seabornfont
    },
    legend image post style={
        xshift=0pt,  
    },
    legend cell align=left,
    legend image code/.code={\draw[fill=#1] (0.2cm, 0.13cm) rectangle ( 0cm, -0.07cm);},
]

\addplot [fill=blue!80,draw=black,bar shift=-20pt] coordinates {
  (SID-CA Test,0.846)
  (CASP-16,0.352)
};

\addplot [fill=blue!45,draw=black,bar shift=-12pt] coordinates {
  (SID-CA Test,0.776)
  (CASP-16,0.326)
};
    
\addplot [fill=blue!10,draw=black,bar shift=-4pt] coordinates {
  (SID-CA Test,0.835)
  (CASP-16,0.329)
};

\addplot [fill=green1,draw=black,bar shift=4pt] coordinates {
  (SID-CA Test,0.352)
  (CASP-16,0.245)
};

\addplot [fill=green4,draw=black,bar shift=12pt] coordinates {
  (SID-CA Test,0.528)
  (CASP-16,0.254)
};

\addplot [fill=green7,draw=black,bar shift=20pt] coordinates {
  (SID-CA Test,0.817)
  (CASP-16,0.102)
};

\legend{
3-hop Interface,
3-hop CDR,
Nondocking,
Random rotation,
Mask embed,
Mask embed +Rotation
}

\end{axis}
\end{tikzpicture}
\caption{Performance comparison of different data augmentation strategies on CASP-16 benchmark in AP score. Here, `interface' and `CDR' mean the selected seed sets for initializing the sampling procedure. `Nondocking' is the setting, where inter Ig-Ag edges are removed from $\mathcal{G}$. For a detailed description of augmentation strategies, please refer to \cref{sec:data_augmentation}. `3-hop' refers to a 3-hop BFS sampling procedure starting from a set of seed nodes and finishing once reaching a node threshold (600). All models perform weighted sum over all nodes in graphs. }
\label{fig:data_augmentation}
\end{figure}
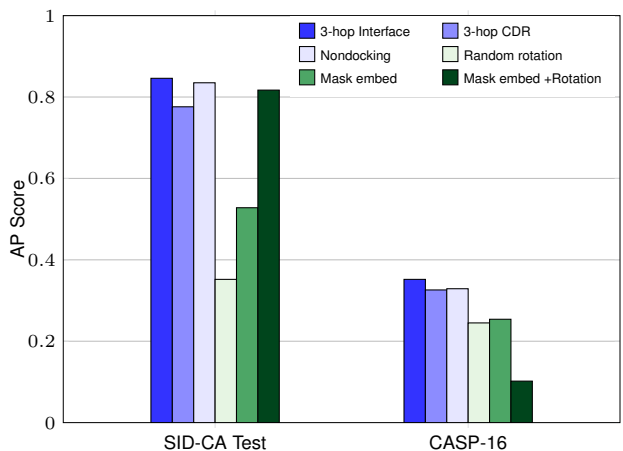

Here, we describe the data augmentation methods presented in \textbf{\cref{fig:data_augmentation}}. A detailed comparison is shown in \textbf{\cref{tab:pooling_strategy}}. The sampling algorithm is described in \textbf{\cref{alg:khop_sampling}}.

\begin{itemize}
    \item \textbf{Nondocking}: $\mathcal{E}_{\text{inter}}$ is discarded from computational graphs when input to models. In this setting, no sampling method is applied to graphs.
    \item \textbf{Random rotation}: During training, positions of nodes in input graphs are rotated by  random angles.
    \item \textbf{Mask embed}: Residues of nodes excluded from the sampling process are masked out from the residue sequence before performing ESM-2.
    \item \textbf{Mask embed + random rotation}: First random rotation is applied to graphs in training. Then, residue characters of unselected nodes in a computational graph are masked out from the input sequence before the embedding generation step.
    \item \textbf{3-hop sampling with node threshold (3-hop interface)}: All nodes included in inter-Ig-Ag edges are selected as seed nodes. A Breadth First Search (BFS) sampling procedure starts picking nodes layer by layer. The process stops when the number of selected nodes exceeds a pre-defined threshold.
    \item \textbf{CDR sampling with node threshold (3-hop CDR)}: Similar to the strategy above but seed nodes are only those located at the CDR region.
\end{itemize}

This experiment investigated whether graph-level augmentation strategies improve model robustness under distribution shifts. We grouped augmentations into three main categories: (i) graph sampling based on $k$-hop iterations with interface or CDR anchors; (ii) geometric perturbations with random rotations in training; and (iii) node embedding modifications with sequence masking before ESM-2 execution. For a detailed view of their results on the five metrics, please refer to \textbf{\cref{tab:data_augmentation}}.

As shown in \textbf{\cref{fig:data_augmentation}} and \textbf{\cref{tab:data_augmentation}}, the 3-hop interface method achieved the highest AP score (0.846) on the SID-CA test set. Although all methods suffer performance drop on CASP-16, the 3-hop interface sampling approach remained the most resilient, maintaining the highest AUC (0.891) and AP (0.352) scores. Methods that applied additional geometric changes or embedding-level noise, such as `Random rotation', `Mask Embed' and `Mask Embed + random rotation', saw a drastic decline in performance on CASP-16, suggesting that perturbing coordinates or features is insufficient for cross-dataset generalization. Moreover, the `Nondocking' approach achieved performance comparable to the CDR-based sampling method, suggesting that the model likely learned shapes from individual protein graphs rather than relying solely on explicit interaction edges. 

\subsection{Graph Construction Methods} 
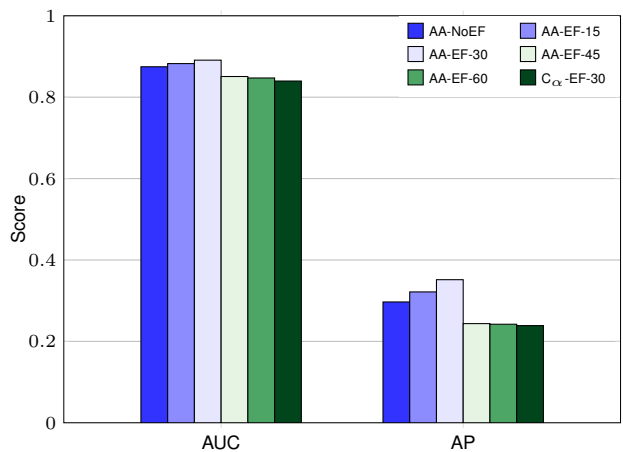
\begin{figure}[ht]
\centering
\begin{tikzpicture}
\begin{axis}[
    width=8.9cm,
    height=7cm,
    ybar,
    bar width=10pt,
    enlarge x limits=0.65,
    enlarge y limits={upper,value=0.0},
    legend style={
        at={(0.80,0.8)},
        anchor=south,
        legend columns=2,
        column sep=0.2cm,
        /tikz/every even column/.append style={column sep=10pt},
        /tikz/every odd column/.append style={column sep=0pt},
        draw=none,
        font=\tiny\seabornfont
    },
    legend image post style={
        xshift=0pt,  
    },
    legend cell align=left,
    legend image code/.code={\draw[fill=#1] (0.2cm, 0.13cm) rectangle ( 0cm, -0.07cm);},
    ymajorgrids=true,
    ytick={0,0.2,0.4,0.6,0.8,1},
    ylabel={Score},
    ylabel shift=-0.15cm,
    symbolic x coords={AUC,AP},
    xtick=data,
    ymax=1.0,
    ymin=0,
    ymajorgrids=true,
    tick label style={font=\small\seabornfont}
]

\addplot [fill=blue!80,draw=black,bar shift=-25pt] coordinates {
  (AUC,0.8747193187)
  (AP,0.2968704005)
};
\addplot [fill=blue!45,draw=black,bar shift=-15pt] coordinates {
  (AUC,0.8825150478)
  (AP,0.321540873)
};

\addplot [fill=blue!10,draw=black,bar shift=-5pt] coordinates {
  (AUC,0.8909621509)
  (AP,0.3515720679)
};

\addplot [fill=green1,draw=black,bar shift=5pt] coordinates {
  (AUC,00.8507688002)
  (AP,0.2436457798)
};

\addplot [fill=green4,draw=black,bar shift=15pt] coordinates {
  (AUC,0.8471611907)
  (AP,0.2421673079)
};
		
\addplot [fill=green7,draw=black,bar shift=25pt] coordinates {
  (AUC,0.8396610851)
  (AP,0.2387877841)
};

\legend{AA-NoEF,AA-EF-15,AA-EF-30,AA-EF-45,AA-EF-60,$\text{C}_\alpha$-EF-30}
\end{axis}
\end{tikzpicture}
\caption{Performance comparison across graph construction methods on CASP-16. Here, AA (all-atom) denotes the baseline setting, $\text{C}_\alpha$ means edges in a graph are established based on $\text{C}_\alpha$ distances, and EF represents edge features with a corresponding size. All models perform weighted sum global pooling over all nodes in graphs.}
\label{fig:graph_construction}
\end{figure}

To assess the impact of graph topology and edge-embedding sizes on CASP-16 \citep{CASP16} performance, we compared various graph construction schemes. 
The all-atom (AA) strategy consistently outperformed the $\text{C}_\alpha$-based method in AUC and AP (\textbf{\cref{fig:graph_construction}}), indicating that side-chain and backbone atoms provide important geometric information. 

In AA models, small edge embedding dimensions (0-30) had no discernible impact on model performances, likely because the EGNN architecture inherently encodes distance information within its radial network (\textbf{\cref{fig:graph_construction}}). However, larger embedding sizes (45 and 60) substantially degraded model performance, resulting in an approximate 10-point drop in AP score. These results highlight the critical need to explore alternative strategies for generating edge attributes to improve model generalization.

\subsection{Overfitting and generalization challenges} 

\begin{figure}[ht]
  \centering
  \includegraphics[width=\columnwidth]{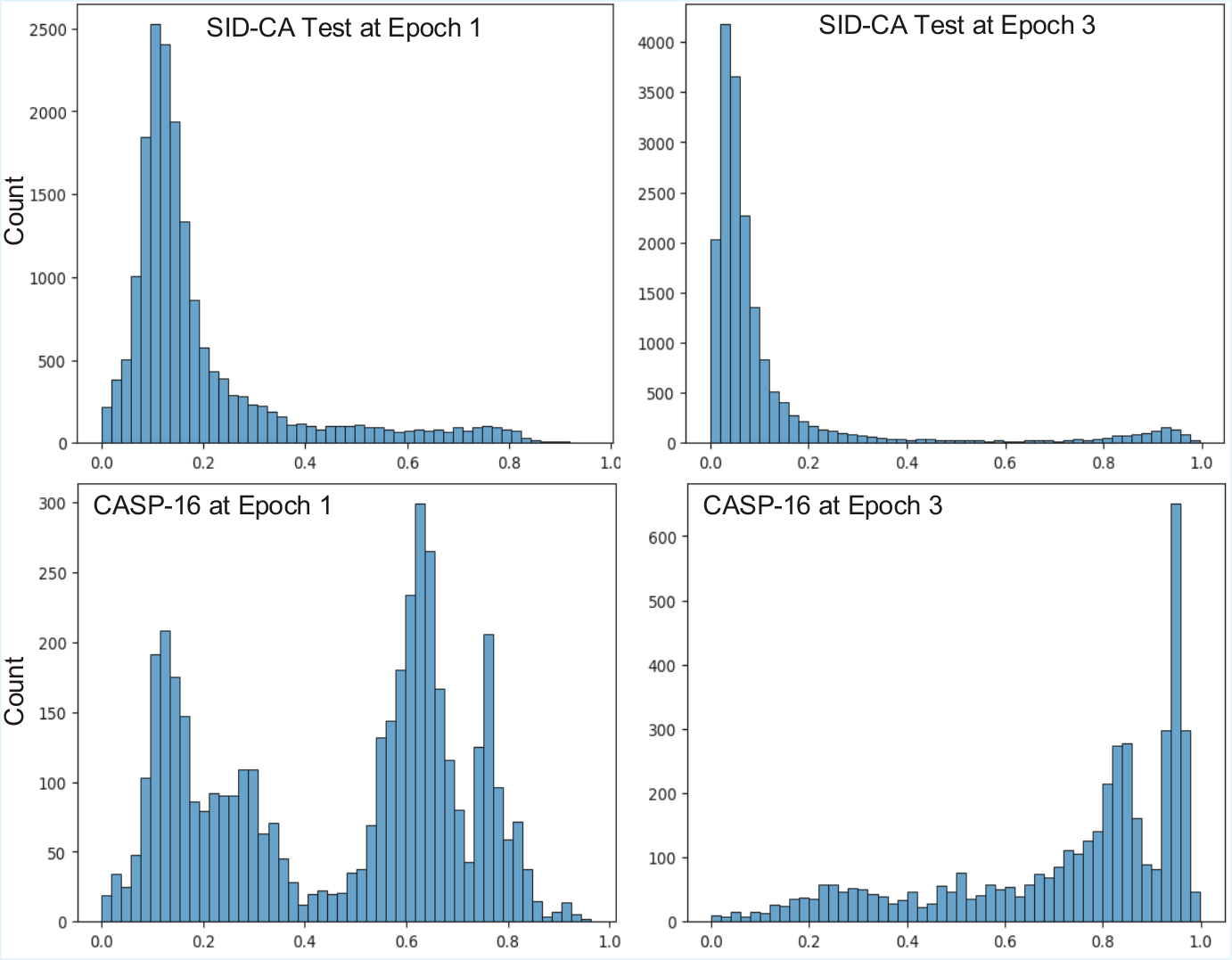}%
    \caption{Change of probability distributions over three epochs for SID-CA test set (top) and CASP-16 (bottom). Probability distributions of the two test datasets change differently as training progresses.}
  \label{fig:all-prob-dists}
\end{figure}

Given the complex contextual nature of Ig-Ag, the distribution shift problem is inevitable.
We performed inferences on the SID-CA test set and CASP-16 \citep{CASP16} and visualized their probability distributions.
The two datasets show different trends in probability distribution as the training process progresses: 
the SID-CA test set's probabilities gradually shift to the left and have a long right tail reflecting the imbalance of positive samples in the training set, while the CASP-16 dataset's probabilities shift to the right indicating model bias toward positive samples (\textbf{\cref{fig:all-prob-dists}}). 
Furthermore, the AP score in the SID-CA test set increases steadily from Epochs 1 to 5 and remains stable, while this score in the CASP-16 benchmark degrades as training progresses (\textbf{\cref{fig:test_set_vs_casp16}}). 
These results indicate that there is a degree of overfitting in the current training paradigm, especially when an oversampling method is used to increase the occurrence of positive samples.
Practically, we can apply ensemble techniques to aggregate predictions of different models to alleviate this problem by leveraging the diversity of individual models to create a more robust and generalized prediction.

\subsection{Customized GRU for Learning Acceleration} 

\begin{figure}[ht]
    \centering
    \begin{tikzpicture}
      \begin{axis}[
        width=9cm,
        height=6.5cm,
        xlabel={\#Epoch},
        ylabel={AUC Score},
        xmin=1, xmax=15,
        ymin=0.55, ymax=1.05,
        grid=major,
        ylabel shift=-0.3cm,
       legend style={
        at={(0.25,0.0)},
        anchor=south,
        legend columns=1,
        draw=none,
        font=\tiny\seabornfont},
        nodes={               
          inner xsep=8pt,
        },
        column sep=0.1cm,
        legend cell align=left,
      ]

        \addplot[color=ForestGreen,mark=*,] table[row sep=\\,y=OurTestSingle,x=Epoch] {
          Epoch  OurTestSingle \\
          1      0.9653695871026574 \\
            2      0.9649114616537684 \\
            3      0.9686136940452077 \\
            4      0.9687137484162787 \\
            5      0.974004178452109 \\
            6      0.9934732018205735 \\
            7      0.9955119278428302 \\
            8      0.9886281618946553 \\
            9      0.9871024430667301 \\
            10     0.993550181270528 \\
            11     0.9841672311406022 \\
            12     0.9891450779554056 \\
            13     0.986465171748025 \\
            14     0.9884655656608861 \\
            15     0.9806515525513044 \\
        };
        \addlegendentry{SID-CA Test (Original GRU)}

        \addplot[color=blue!80,mark=diamond*] table[row sep=\\,y=CASP-16,x=Epoch] {
          Epoch  CASP-16  \\
          1      0.7469219002653454 \\
          2      0.8668849365840949 \\
          3      0.8054729010548678 \\
          4      0.7535530300698545 \\
          5      0.7674552108841005 \\
          6      0.7412799998854727 \\
          7      0.7283455746299158 \\
          8      0.71967514335595 \\
          9      0.7330741199114992 \\
          10     0.7186311555731483 \\
          11     0.7127777376136951 \\
          12     0.6628753363344254 \\
          13     0.6922608902921662 \\
          14     0.6433276761626489 \\
          15     0.6736523538578159 \\
        };
        \addlegendentry{CASP-16 (Original GRU)}
        
        \addplot[color=DeepBlue,mark=square*] table[row sep=\\,y=OurTestDouble,x=Epoch] {
          Epoch  OurTestDouble  \\
          1      0.9655802245656367 \\
          2      0.967860764412873 \\
          3      0.9691041306882833 \\
          4      0.9694996197051252 \\
          5      0.97080827035435 \\
          6      0.971712164190547 \\
          7      0.9934936918461551 \\
          8      0.9953915095386429 \\
          9      0.9956094603645989 \\
          10     0.9952696411711065 \\
          11     0.9929543943728464 \\
          12     0.9917324953396226 \\
          13     0.9949946335046845 \\
          14     0.9957529535899029 \\
          15     0.9906414802851911 \\
        };
        \addlegendentry{SID-CA Test (Customized GRU)}

        \addplot[color=MaroonBrickRed,mark=triangle*] table[row sep=\\,y=CASP-16,x=Epoch] {
          Epoch  CASP-16  \\
          1      0.8794450007766381 \\
          2      0.8613189105305692 \\
          3      0.8132274719461836 \\
          4      0.7664720657214825 \\
          5      0.7366140867129023 \\
          6      0.7357884165672307 \\
          7      0.7318651412586692 \\
          8      0.7163910734570847 \\
          9      0.7303648338244885 \\
          10     0.7254201182923695 \\
          11     0.7345372059780381 \\
          12     0.7184525645880203 \\
          13     0.7270929324496599 \\
          14     0.6605060531249127 \\
          15     0.7156044141678841 \\
        };
        \addlegendentry{CASP-16 (Customized GRU)}

      \end{axis}
    \end{tikzpicture}
    \caption{AUC Score trends over 15 training epochs for IgPoseClassifier's GRU variants on SID-CA test set and CASP-16. All models use the all-node global pooling operator in this experiment. }
    \label{fig:test_set_vs_casp16}
\end{figure}
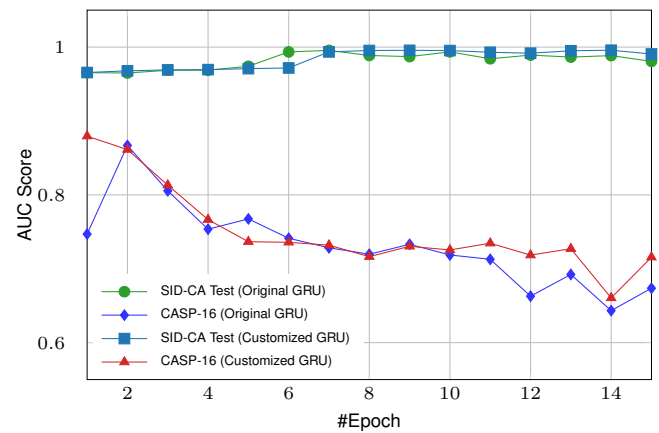

We customized the original GRU cell \citep{cho2014learning} by adding an additional connection between the previous hidden state and the current input. Practically, this modification accelerates model learning. As shown in \textbf{\cref{fig:test_set_vs_casp16}}, AP scores of the customized version are higher than those of the original version in the first epoch on both datasets, especially on CASP-16 \citep{CASP16}. The trend in performance gain continues until Epoch 5 on our test set. These results indicate that the additional connection in GRU enables models to learn quickly from the training data.

\subsection{Performance of Top EMA Methods on CASP-16}

\begin{figure}[ht]
\centering
\begin{tikzpicture}
\begin{axis}[
    width=8.9cm,
    height=7cm,
    ybar,
    bar width=8pt,
    enlarge x limits=0.3,
    enlarge y limits={upper,value=0},
    legend style={
        at={(0.83,0.68)},
        anchor=south,
        legend columns=1,
        column sep=0.2cm,
        /tikz/every even column/.append style={column sep=10pt},
        /tikz/every odd column/.append style={column sep=0pt},
        draw=none,
        font=\tiny\seabornfont
    },
    legend image post style={
        xshift=0pt,  
    },
    legend cell align=left,
    legend image code/.code={\draw[fill=#1] (0.2cm, 0.13cm) rectangle ( 0cm, -0.07cm);},
    ymajorgrids=true,
    ytick={0,0.2,0.4,0.6,0.8,1},
    ylabel={Score},
    ylabel shift=-0.15cm,
    symbolic x coords={AUC,AP,r},
    xtick=data,
    ymax=1.0,
    ymin=0,
    tick label style={font=\small\seabornfont}
]

  \addplot+[bar shift=-16pt,draw=black, fill=blue!70!black,error bars/.cd,y dir=both,y explicit,error bar style={draw=black, line width=0.5pt},
    error mark options={draw=black, line width=3pt, mark size=0.1pt}] coordinates {
      (AUC,0.914)
      (AP,0.747)
      (r,0.3547)
    };

\addplot [fill=blue!80,draw=black,bar shift=-8pt] coordinates {
  (AUC,0.8940)
  (AP,0.3261)
  (r,0.2316)
};

\addplot [fill=blue!15,draw=black,bar shift=0pt] coordinates {
  (AUC,0.6984)
  (AP,0.1163)
  (r,0.0450)
};

\addplot [fill=green1,draw=black,bar shift=8pt] coordinates {
  (AUC,0.6986)
  (AP,0.1192)
  (r,0.1915)
};

\addplot [fill=green7,draw=black,bar shift=16pt] coordinates {
  (AUC,0.8604)
  (AP,0.2640)
  (r,0.3366)
};

\legend{IgPose,MIEnsembles-Server,PIEFold\_human,MULTICOM\_AI,MULTICOM\_GATE}
\end{axis}
\end{tikzpicture}
\caption{Performance of IgPose and top EMA methods reported in CASP-16 \citep{CASP16}. We downloaded predicted results from CASP-16 and computed AUC, AP, and $r$ scores.}
\label{fig:casp16_top_methods}
\end{figure}
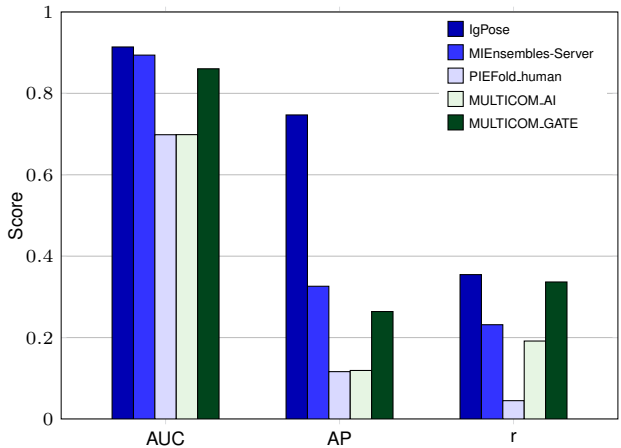

In the Quality Assessment (QA) category of CASP-16 \citep{CASP16}, participants do not predict structures; rather, they evaluate and rank ensembles of decoys generated by other prediction servers. To benchmark IgPose against these state-of-the-art Estimation of Model Accuracy (EMA) methods, we retrieved all submissions corresponding to our eight CASP-16 structural targets.
We then extracted both the overall score and the interface scores from 'QMODE 1' in all submissions. Due to missing interface scores in several entries, we evaluated performance using AUC, AP, and Pearson correlation ($r$) based on overall scores and DockQ. 
For clarity, we report only the top four performing EMA methods\textemdash prioritized by Area Under the Precision-Recall Curve (AP)\textemdash as the remaining entries yielded AP scores below 0.1. As illustrated in \textbf{\cref{fig:casp16_top_methods}}, the predictive performance of current EMA methods remains limited on Ig–Ag targets, particularly in terms of AP and $r$ scores. These results underscore the inherent difficulty of the antibody–antigen binding prediction problem and suggest that general-purpose QA methods struggle to capture the specific biophysical nuances of immunoglobulin interfaces.

\subsection{Analysis of Score Distributions of Rosetta and Prodigy}

\begin{figure*}[htbp]
    \centering
    \subfloat[Rosetta Binding Energy]{
        \includegraphics[width=0.43\linewidth]{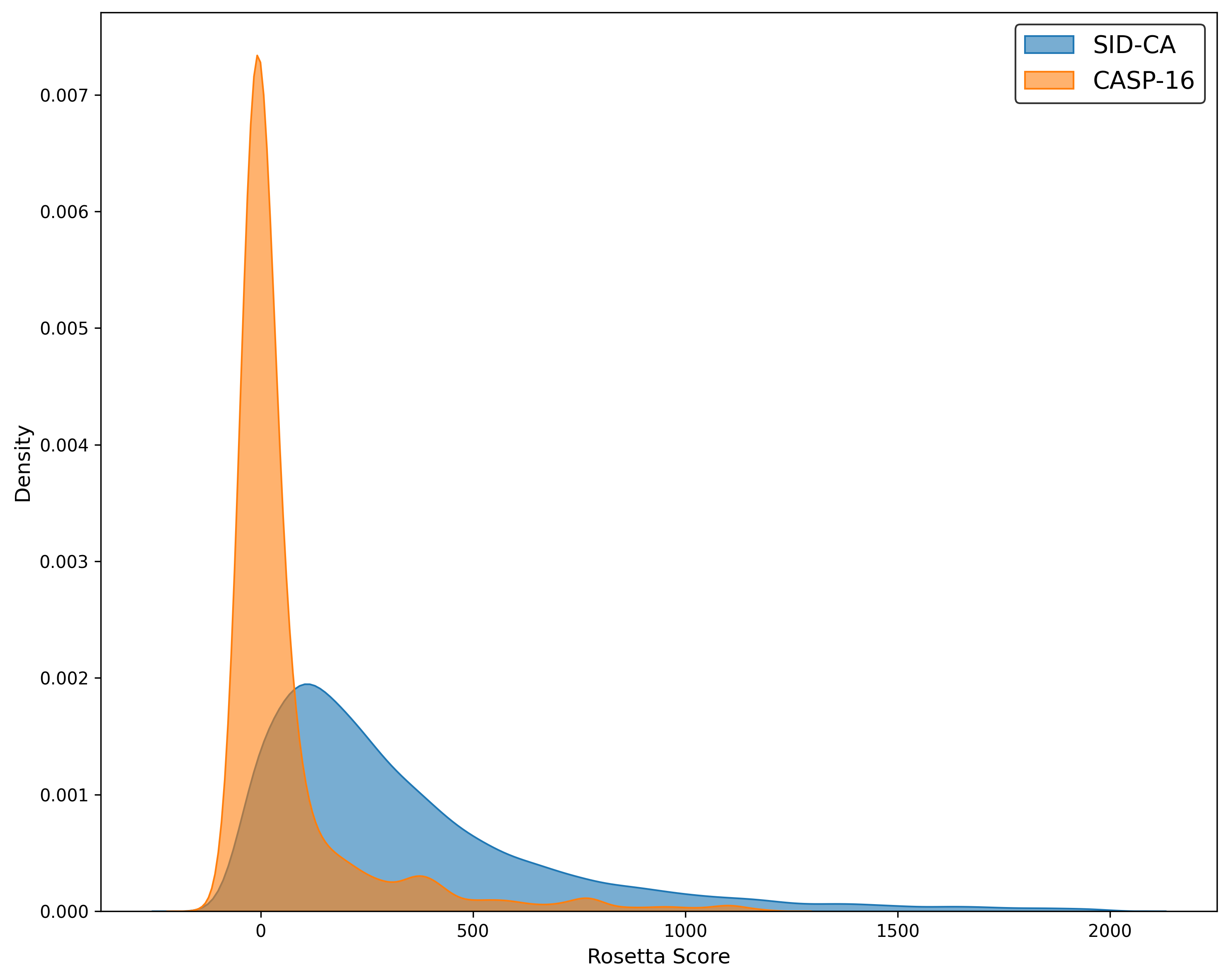}
    }
    \hfil
    \subfloat[Prodigy Binding Affinity]{
        \includegraphics[width=0.43\linewidth]{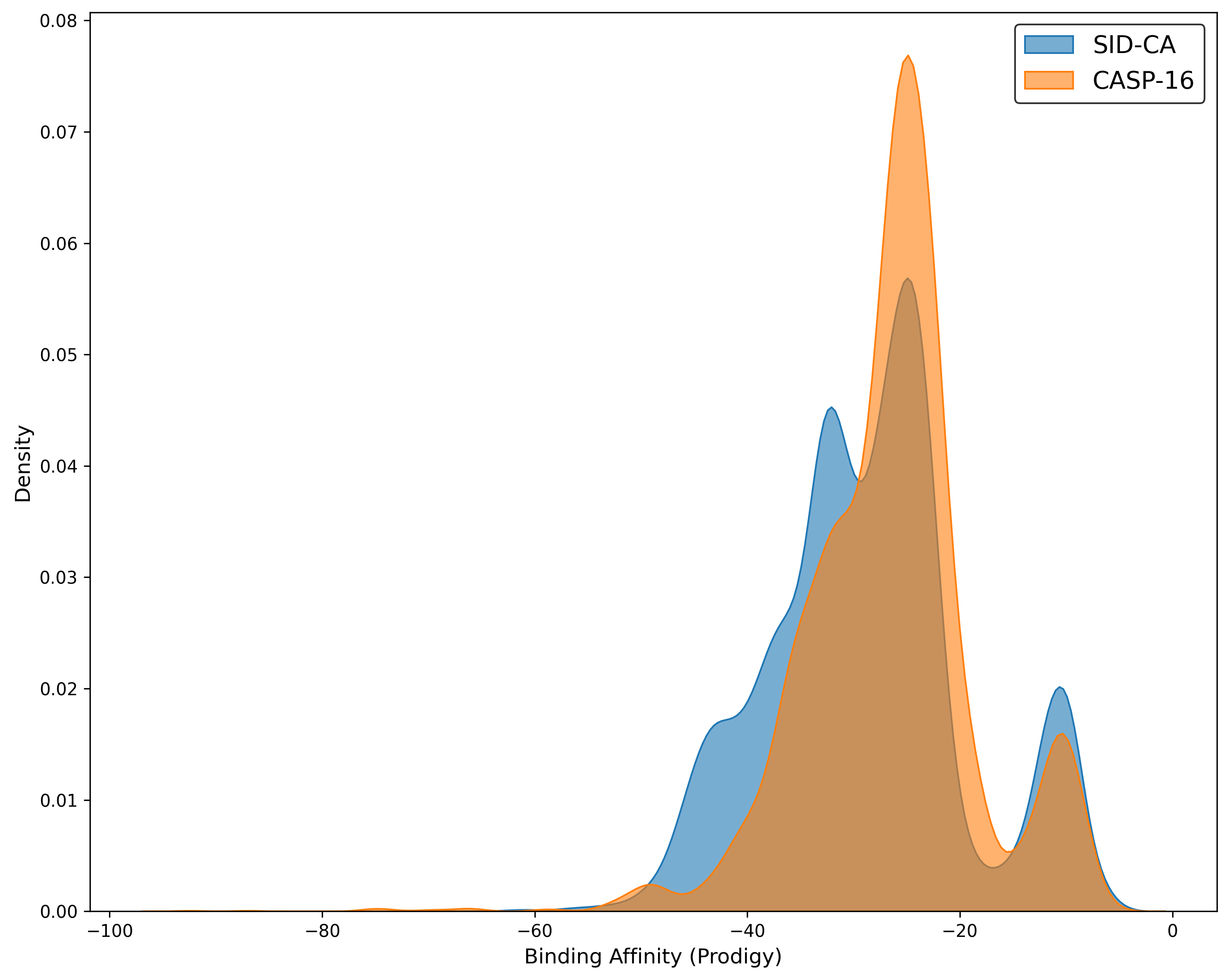}
    }
    \caption{Density distributions of Rosetta binding energy and Prodigy binding affinity scores. CASP-16's structures show narrower distributions concentrated at lower energy values compared to SID-CA, reducing the discriminative power of energy-based scoring functions.}
    \label{fig:rosetta_prodigy_score}
\end{figure*}

We investigated the discrepancies in the predictive performance of physics-based energy estimation baselines, specifically Rosetta \citep{alford2017rosetta} and Prodigy \citep{xue2016prodigy} (\textbf{\cref{fig:rosetta_prodigy_score}}). 
For Rosetta, the CASP-16 structures exhibited a significantly narrower score distribution concentrated at highly favorable (low) binding energies compared to the broader distribution observed in the SID-CA dataset. We attribute this to the fact that CASP-16 submissions typically undergo extensive conformational refinement and energy minimization. This process effectively flattens the local energy landscape, diminishing Rosetta’s discriminative power as both near-native and incorrect poses occupy a similar low-energy range. This phenomenon explains the decline in Rosetta's classification performance on the CASP-16 benchmark.
In contrast, Prodigy exhibited an inverse predictive pattern on the SID-CA dataset (AUC = 0.071; \textbf{\cref{tab:detailed_classification}}). This is likely because Prodigy relies on a contact-counting heuristic, causing synthetic decoys with high-density, hallucinated interfacial contacts to receive more favorable binding affinity scores than the true native poses. 
On CASP-16, Prodigy achieved moderate performance (AUC = 0.677, $r = 0.381$, AP = 0.129). This suggests that while Prodigy can differentiate between tight and loose interfacial packing, it lacks the geometric sensitivity required to distinguish high-quality complexes (DockQ $\ge 0.8$) from `acceptable' to `medium' quality decoys \citep{basu2016dockq}.

\subsection{Classification Performance on Ab-Ag, Nb-Ag, TCR-pMHC subsets}

\begin{figure}[ht]
\centering
\begin{tikzpicture}
\begin{axis}[
    width=8.9cm,
    height=7cm,
    ybar,
    bar width=10pt,
    enlarge x limits=0.3,
    enlarge y limits={upper,value=0},
    legend style={
        at={(0.87,0.73)},
        anchor=south,
        legend columns=1,
        column sep=0.2cm,
        /tikz/every even column/.append style={column sep=10pt},
        /tikz/every odd column/.append style={column sep=0pt},
        draw=none,
        font=\tiny\seabornfont
    },
    legend image post style={
        xshift=0pt,  
    },
    legend cell align=left,
    legend image code/.code={\draw[fill=#1] (0.2cm, 0.13cm) rectangle ( 0cm, -0.07cm);},
    ymajorgrids=true,
    ytick={0,0.2,0.4,0.6,0.8,1},
    ylabel={Score},
    ylabel shift=-0.15cm,
    symbolic x coords={Ab, Nb, TCR},
    xtick=data,
    ymax=1.0,
    ymin=0,
    tick label style={font=\small\seabornfont}
]

\addplot [fill=blue!80,draw=black,bar shift=-15pt] coordinates {
    (Ab, 0.9449)
    (Nb, 0.9984)
    (TCR, 0.2011)
};

\addplot [fill=blue!15,draw=black,bar shift=-5pt] coordinates {
    (Ab, 0.8660)
    (Nb, 0.9970)
    (TCR, 0.2281)
};

\addplot [fill=green1,draw=black,bar shift=5pt] coordinates {
    (Ab, 0.9412)
    (Nb, 0.8647)
    (TCR, 0.0070)
};

\addplot [fill=green7,draw=black,bar shift=15pt] coordinates {
    (Ab, 0.9680)
    (Nb, 0.8143)
    (TCR, 0.0094)
};

\legend{SID-CA AUC, SID-CB AUC, SID-CA AP, SID-CB AP}

\end{axis}
\end{tikzpicture}
\caption{Performance comparison on Ab-Ag, Nb-Ag, and TCR-pMHC subsets. }
\label{fig:target_fixed}
\end{figure}
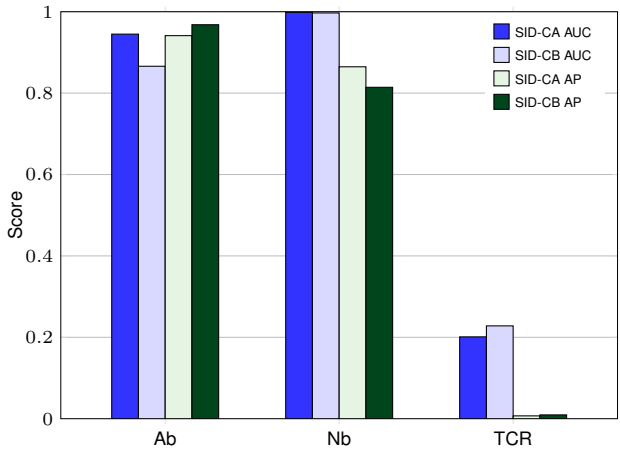

 We evaluated the performance of IgPoseClassifier across the Ab-Ag, Nb-Ag, and TCR-pMHC subsets, with dataset statistics (\textbf{\cref{tab:sid_stats}}) and classification metrics (\textbf{\cref{fig:target_fixed}}).

 As shown in \textbf{\cref{fig:target_fixed}}, IgPose demonstrates high predictive accuracy for Ab-Ag and Nb-Ag subsets, but encounters specific challenges when evaluating TCR-pMHC binding interactions. While AUC and AP scores remain high for Ab-Ag and Nb-Ag, the lower evaluation scores for TCR-pMHC suggest that this subset is more difficult for the current architecture to characterize accurately.
The difference in performance is likely influenced by the unique biophysical properties of TCR-pMHC pairings, which are often of lower affinity and structurally more rigid compared to the more flexible Ab-Ag interactions. Furthermore, we observed an extreme data imbalance within the TCR subset, which contains the smallest ratio of positive samples (\textbf{\cref{tab:sid_stats}}), presenting a huge challenge for IgPose in learning distinctive geometric patterns. 
These results suggest two potential directions for improving IgPose's performance on TCR-pMHC subset: (i) investigating the geometric differences between TCR-pMHC and Ab/Nb-Ag binding interfaces, and (ii) addressing the severe data imbalance in this subset.

\begin{table}[ht]
\centering
\caption{Dataset Statistics for Ab-Ag, Nb-Ag, TCR-pMHC Subsets in SID-CA and SID-CB.}
\begin{tabular}{c|cc|cc}
\toprule
\multirow{2}{*}{\textbf{Type}} & \multicolumn{2}{c|}{\textbf{SID-CA}} & \multicolumn{2}{c}{\textbf{SID-CB}} \\ \cline{2-5} 
 & \multicolumn{1}{c}{\textbf{\#pos}} & \multicolumn{1}{c|}{\textbf{\#neg}} & \multicolumn{1}{c}{\textbf{\#pos}} & \multicolumn{1}{c}{\textbf{\#neg}} \\ \midrule
Ab-Ag & 701 & 679 & 738 & 153 \\
Nb-Ag & 198 & 14,230 & 228 & 14,126 \\
TCR-pMHC & 27 & 2,139 & 35 & 2,157 \\ 
\bottomrule
\end{tabular}
\label{tab:sid_stats}
\end{table}

\section{Algorithms}
This section presents fundamental algorithms embedded in the IgPose architecture.

\subsection{Theoretical Analysis of Global Pooling and Information Propagation}

 To interpret the ablation results regarding the read-out set $\mathcal{S}$, we analyze the information flow within the network. Let the input graph be $\mathcal{G} = (\mathcal{V}, \mathcal{E}, \mathcal{X}_v, \mathcal{X}_e, \mathcal{P})$, where $\mathcal{V}$ is the set of residues. The network depth is defined by $T$ layers. 

 As described in \textbf{\cref{eq:egnn}}, the node features $H^{(t)}$ at layer $t$ are updated via an EGNN layer followed by a customized GRU gated update. We can rewrite \textbf{\cref{eq:egnn}} in the node-level format as follows: 

\begin{equation}
\begin{aligned}
\tilde{h}_i^{(t)}, p_i^{(t)}  & = \text{EGNN} \left(h_i^{(t-1)}, p_i^{(t-1)}, \{h_j^{(t-1)}, p_j^{(t-1)}\}_{j \in \mathcal{N}(i)}\right),  \\
h_i^{(t)} & = \text{GRU}\left( \bigl[ \tilde{h}_i^{(t)}, h_i^{(t-1)} \bigr], h_i^{(t-1)} \right) 
\end{aligned}
\label{eq:egnn_node}
\end{equation}

\noindent  where $i$ and $j$ are node indices, and $\mathcal{N}_i$ denotes a set of neighbors of the node $i$. 

 \textbf{\cref{eq:egnn_node}} establishes that $h_i^{(t)}$ is a function of the local neighborhood at $t-1$. We define the structural receptive field $\mathcal{R}_i^{(t)}$ of node $i$ at the layer $t$ as the set of input nodes that influence its state: 

\begin{equation}
     \mathcal{R}_i^{(0)} = {i}, \quad \mathcal{R}_i^{(t)} = \mathcal{R}_i^{(t-1)} \cup \bigcup_{j \in \mathcal{N}(i)} \mathcal{R}_j^{(t-1)}. 
\end{equation}

\noindent  In other words, a node $i$'s state at the layer $t$ depends on every node that directly influenced $i$ and other nodes that affected each of its neighbors at the previous layer $t-1$. 

Given $T$ layers, the final embedding $h_i^{(T)}$ aggregates information from the $T$-hop neighborhood of node $i$. Since the protein graph is connected via peptide bonds and inter-residue contacts, for a sufficient $T$, the receptive field of a framework node $v_{\text{frame}}$ expands to include interface nodes $v_{\text{int}}$. Thus, the final embedding $h^{(T)}$ of a non-interface node $i$ is conditionally dependent on the state of interface nodes:

\begin{equation}
 h_i^{(T)} = f \bigl( \bigl\{ h_k^{(0)} : k \in \mathcal{R}_i^{(T)} \bigr\} \bigr) 
\end{equation}

The global graph embedding $g$ is computed via a weighted sum over a selected subset $\mathcal{S} \subset \mathcal{V}$:

\begin{equation}
 g(\mathcal{S}) = \sum_{i \in \mathcal{S}} \sigma(w_p^\top h_i^{(T)} + b_p) \odot h_i^{(T)}. 
\end{equation}

Specifying a subset $\mathcal{S} $ that maximizes classification performance (perturbation method) is prevalent in interpretable GNNs \citep{bui2023generating, bui2023toward}. The exploration procedure can be guided by either learning algorithms or domain expertise. Given the large scale of computational protein graphs, we opt for the latter approach and left the first one for future exploration. Our empirical observations demonstrate that defining $\mathcal{S}$ as the set of non-interface (or non-CDR-epitope) nodes maximizes discriminative power (high AP \& AUC scores). This phenomenon can be referred to as the \emph{inductive bias} of message-passing networks in decoy discrimination tasks.

While the interface region ($\mathcal{V}_{\text{int}}$) contains the direct binding contacts, it is also the region of highest variance and noise in generated decoys (e.g., local steric clashes or side-chain overlap). Conversely, the connecting framework nodes ($\mathcal{V}_{\text{frame}}$) can act as specific ``sensors" that integrate these local perturbations. A ``bad" interface induces gradient updates that propagate to the framework, manifesting as geometric strain or latent feature inconsistency in the surrounding nodes.

Therefore, by performing weighted sum pooling over $\mathcal{S} = \mathcal{V} \setminus \mathcal{V}_{\text{int}}$, the readout function focuses on the propagated structural consistency of the complex rather than the noisy local features of the contact boundary. The model learns to discriminate true poses versus wrong ones through indirect signals passed to ``scaffold'' regions.

\subsection{Mathematical Justification for Modified GRU}

The standard GRU \cite{cho2014learning} updates the hidden state $H^{(t-1)}$ through a gating mechanism. Formally, the reset gate $r^{(t)}$ and update gate $z^{(t)}$ are computed as: 

\begin{equation}
\begin{aligned}
    r^{(t)} &= \sigma \left( W_{r}^{(i)} \widetilde{H}^{(t)} + W_{r}^{(h)} H^{(t-1)} + b_r \right)
    \\
    z^{(t)} &= \sigma \left( W_{r}^{(i)} \widetilde{H}^{(t)} + W_{r}^{(h)} H^{(t-1)} + b_z \right)
\end{aligned}
\end{equation}

In the candidate hidden state $n^{(t)}$, the reset state $r^{(t)}$ multiplies to $H^{(t-1)}$ in element-wise, strictly controlling the historical context used for the new proposal: 

\begin{equation}
     n^{(t)} = \tanh \left( W_{n}^{(i)} \widetilde{H}^{(t)} + W_{n}^{(h)} (r^{(t)} \odot H^{(t-1)}) + b_n \right) 
\end{equation}

The final hidden state $H^{(t)}$ is a linear interpolation between the previous state and the candidate state:
\begin{equation}
    H^{(t)} = (1 - z^{(t)}) \odot n^{(t)} + z^{(t)} \odot H^{(t-1)}
\end{equation}

We modify the GRU function by defining an input vector $X^{(t)} = \bigl[\widetilde{H}^{(t)}, H^{(t-1)} \bigr]$. By expanding the update gate $z^{(t)}$ (or the reset gate $r^{(t)}$), we can group terms as follows:

\begin{equation}
\begin{aligned}
z^{(t)} &= \sigma \left( W_{z}^{(i)} X^{(t)} + W_{z}^{(h)} H^{(t-1)} + b_z \right) \\
        &= \sigma \left( W_z^{(i,\widetilde{H})} \widetilde{H}^{(t)} + \mathbf{W_z^{(i, H)} H^{(t-1)}} + W_z^{(h)} H^{(t-1)} + b_z \right) \\
        &= \sigma \left( W_z{(i,\widetilde{H})} \widetilde{H}^{(t)} + (\mathbf{W_z^{(i, H)}} + W_z^{(h)}) H^{(t-1)} + b_z \right) 
\end{aligned}
\end{equation}

As can be seen, we are now having two independent weights operating on $H^{(t-1)}$. If we use Xavier or Kaiming method for weight initialization, these operations can increase pre-activation variance and saturation of $r$ and $z$ gates. Specifically, increasing the magnitude of pre-activation via the double-weight transformation pushes the gate values away from 0.5 toward the saturation regions (0 or 1) as the Sigmoid function is most sensitive in range $[-2, 2]$. Furthermore, $r$ and $z$ gates are more sensitive to $H^{(t-1)}$ as a node is listening to its own history twice as loudly as it is listening to the neighbors. 

In standard GNN architectures, node representations tend to become homogeneous across nodes (oversmoothing) as the number of layers increases. The modified GRU enforces a strong self-loop for each node. The candidate state $n^{(t)}$ relies not only on the neighbor-averaged signal $\widetilde{H}^{(t)}$ but also directly on the node's previous features via $W_i^{(n, H)}$. This additional linear transformation allows the model to selectively balance graph topology information with temporal changes, resulting in more stable performance for dynamic physical systems. The modified candidate hidden state $n^{(t)}$ is as follows:

\begin{equation}
    \begin{aligned}
    n^{(t)} = \tanh \bigl( &  W_i^{(n)} X^{(t)} + W_h^{(n)} (r^{(t)} \odot H^{(t-1)}) + b_n \bigr) \\
                  =  \tanh \bigl( &  W_i^{(n, \widetilde{H})} \widetilde{H}^{(t)} + \mathbf{W_i^{(n, H)} H^{(t-1)}}  \\ 
                                  &  \quad \quad + \; W_h^{(n)} (r^{(t)} \odot H^{(t-1)}) + b_n \bigr).
    \end{aligned}
\end{equation}

In a standard GRU, the gradient of the new state w.r.t the old state relies heavily on the active gates. If $z^{(t)}$ and $r^{(t)}$ saturate to 0, the gradient signal diminishes. In contrast, the term $\mathbf{W_i^{(n,H)} H^{(t-1)}}$ in our modified GRU acts as a direct gradient shortcut. Omitting the bounded derivative of the tanh function and considering $r^{(t)}$ as a constant factor, the partial derivative of $n^{(t)}$ w.r.t $H^{(t-1)}$ has a path independent from the reset gate $r^{(t)}$: 

\begin{equation}
\frac{\partial n^{(t)}}{\partial H^{(t-1)}} \propto W_h^{(n)} \cdot \text{diag}(r^{(t)}) + \mathbf{W_i^{(n, H)}}  
\end{equation}

\noindent Therefore, the gradient flow to $H^{(t-1)}$ always sustains even if the reset gate $r^{(t)} \to 0$, facilitating learning over multiple EGNN layers.

\subsection{Interface-focused K-hop sampling}
The interface-focused K-hop sampling algorithm outputs a subgraph centered around the interface region of a given input graph. This iterative algorithm starts from a set of seed nodes, which can be either CDR nodes or any nodes in inter-Ig-Ag edges. The output subgraph includes all sampled nodes and any edges established between them. 

\begin{algorithm}[ht]
\caption{K-hop Subgraph Sampling}
\begin{algorithmic}[1]
\Require $\mathcal{G}$, $k$, node threshold $N_{\max}$, optional seed set $S$
\If{$S = \emptyset$}
  \State $S \gets \{u,v \mid (u,v)\in \mathcal{E}_{\text{inter}}\}$  \Comment{interface seeds}
\EndIf
\State $C\gets S$  \Comment{current selected nodes; $|S|\!<N_{\max}$ by assumption}
\For{$i\gets 1$ to $k$}
  \State $L\gets \text{unique}(\text{BFS\_layer}(\mathcal{G},\,S,\,i))$
  \State $L_{\text{new}}\gets L\setminus C$
  \If{$L_{\text{new}}=\emptyset$ or $|C|+|L_{\text{new}}| > N_{\max}$}
    \State \textbf{break}
  \EndIf
  \State $C\gets C\cup L_{\text{new}}$
\EndFor
\State \Return $\text{subgraph}(\mathcal{G},\,C)$
\end{algorithmic}
\label{alg:khop_sampling}
\end{algorithm}

\subsection{Threshold Selection}
Baseline methods output various ranges of continuous values with different meaning. We acknowledge that a robust classifier must have a unique threshold for all datasets. Therefore, we design \textbf{\cref{alg:f1_threshold}} to select a classification threshold that maximize the F-beta score on the evaluation set. In practical virtual screening tasks, we can also opt for the Top-k thresholding approach to prioritize the most promising leads, ensuring high precision among the top-ranked candidates which are most likely to undergo further lead optimization.

\begin{algorithm}[ht]
\caption{Select threshold maximizing F-beta score based on the evaluation set}
\begin{algorithmic}[1]
\Require $S = \{s_i\}_{i=1}^n$, $s_i \in \mathbb{R}$, $Y = \{y_i\}_{i=1}^n$, $y_i\in\{0,1\}, \beta = 0.25$
\State $\mathcal{T} \gets \mathrm{unique}(S)\,\cup\{0,1\}$ \Comment{Candidate thresholds}
\State $\mathrm{score}^* \gets 0$, \,$\tau^* \gets 0$
\ForAll{$\tau \in \mathcal{T}$}
  \State Predict $\hat{y}_i \gets \mathbf{1}\{s_i \ge \tau\}$ for $i=1,\dots,n$
\State $\mathrm{score} \;\gets\; (1+\beta^2)\cdot\dfrac{P\,\times\,R}{\beta^2} \times P+R$
  \If{$\mathrm{score} > \mathrm{score}^*$}
    \State $\mathrm{score}^* \gets \mathrm{score}$,\quad $\tau^* \gets \tau$
  \EndIf
\EndFor
\State \Return $(\tau^*,\,\mathrm{score}^*)$
\end{algorithmic}
\label{alg:f1_threshold}
\end{algorithm}
\end{document}